\pdfoutput=1
\documentclass[twocolumn,preprintnumbers,amsmath,amssymb,letter]{revtex4}
\usepackage{graphicx}
\usepackage{dcolumn}
\usepackage{bm}
\newcommand{\vpi}{\mathbf{\pi}}
\newcommand{\rar}{\rightarrow}
\newcommand{\ra}{\rangle}
\newcommand{\la}{\langle}
\newcommand{\ttt}{\texttt}
\newcommand{\mb}{\mathbb}
\newtheorem{definition}{Definition}
\newtheorem{theorem}{Theorem}

\newcommand{\qed}{\hfill $\Box$ \hfill \\}

\begin{document}

\preprint{LAUR-06-7791}

\title{Grammar-Based Random Walkers in Semantic Networks\footnote{Rodriguez, M.A., "Grammar-Based Random Walkers in Semantic Networks", Knowledge-Based Systems, volume 21, issue 7, pages 727-739, ISSN: 0950-7051, Elsevier, doi:10.1016/j.knosys.2008.03.030, LA-UR-06-7791, October 2008.}}

\author{Marko A. Rodriguez}
%\email{marko@lanl.gov}
\affiliation{Digital Library Research and Prototyping Team \\
		Los Alamos National Laboratory \\
		Los Alamos, New Mexico 87545 }
		
\date{\today}

 \begin{abstract}
Semantic networks qualify the meaning of an edge relating any two vertices. Determining which vertices are most ``central" in a semantic network is difficult because one relationship type may be deemed subjectively more important than another.  For this reason, research into semantic network metrics has focused primarily on context-based rankings (i.e.~user prescribed contexts). Moreover, many of the current semantic network metrics rank semantic associations (i.e.~directed paths between two vertices) and not the vertices themselves. This article presents a framework for calculating semantically meaningful primary eigenvector-based metrics such as eigenvector centrality and PageRank in semantic networks using a modified version of the random walker model of Markov chain analysis. Random walkers, in the context of this article, are constrained by a grammar, where the grammar is a user defined data structure that determines the meaning of the final vertex ranking. The ideas in this article are presented within the context of the Resource Description Framework (RDF) of the Semantic Web initiative.
\end{abstract}

\maketitle{}

\section{Introduction}

There exists a large collection of centrality metrics that have been used extensively to rank vertices in single-relational (or unlabeled) networks. Any metric for determining the centrality of a vertex in a single-relational network can be generally defined by the function $f : G \rightarrow \mb{R}^{|V|}$, where a single-relational network is denoted $G^1 = (V = \{i,\ldots, j\},E \subseteq V \times V)$ and the range of $f$ is the rank vector representing the centrality value assigned to each vertex in $V$ \footnote{The superscript $1$ on $G^1$ denotes that the network is a single-relational network as opposed to a semantic network which will be denoted as $G^n$.}. The work in \cite{socialanal:wasserman1994,netanal:brandes2005,link:getoor2005} provide reviews of the many popular centrality measures that are currently used today to analyze single-relational networks.

Of particular importance to this article are those metrics that use the primary eigenvector of the network to rank the vertices in $V$ (namely eigenvector centrality \cite{power:bonacich1987} and PageRank \cite{page98pagerank}). If $\mathbf{A} \in \mathbb{R}^{|V| \times |V|}$ is the adjacency matrix representation of $G^1$, then the primary eigenvector of $\mathbf{A}$ is $\vpi$ when $\mathbf{A}\vpi = \lambda\vpi$, where $\lambda$ is the greatest eigenvalue of all eigenvectors of $\mathbf{A}$ and $\vpi \in \mathbb{R}^{|V|}$ \cite{numeri:trefethen1997}. The primary eigenvector has been applied extensively to ranking vertices in all types of networks such as social networks \cite{power:bonacich1987}, scholarly networks of articles \cite{findgem:chen2006} and journals \cite{journalstatus:bollen2006}, and technological networks such as the web citation network \cite{page98pagerank}. In single-relational networks, determining the primary eigenvector of the network can be computed using the power method which simulates the behavior of a collection of random walkers traversing the network \cite{netanal:brandes2005}. Those vertices that have a higher probability of being traversed by a random walker are the most ``central" or ``important" vertices. For aperiodic, strongly connected networks, $\vpi$ is the eigenvector centrality ranking \cite{power:bonacich1987}. For networks that are not strongly connected or are periodic, the network's topology can be altered such that a ``teleportation" network can be overlaid with $G^1$ to produce an irreducible and aperiodic network for which the power method will yield a real valued $\vpi$. This is the method that was introduced by Brin and Page and is popularly known as the random web-surfer model of the PageRank algorithm \cite{page98pagerank}. The PageRank algorithm is one of the primary reasons for the (subjectively) successful rankings of web pages from the Google search engine \cite{google:langville2006}.

In a single-relational social network, for example, the network data structure can only represent a single type of relationship such as friendship. However, in a semantic network (or multi-relational network), the vertices can be connected to each other by a heterogeneous set of relationships such as friendship, kinship, collaboration, communication, etc. For a semantic network instance, there usually exists an ontology (or schema) which specifies how vertex types are related to one another. For example, an ontology may say that a vertex of type human can have another vertex of type human as a friend, but a human cannot have a vertex of type animal as a friend. An ontology is nearly analogous to the object-specifications of object-oriented programming minus the method declarations \cite{rodriguez:gpsemnet2007} and loosely related to the schema definitions of relational databases. 

The Resource Description Framework (RDF) is a popular data model for explicitly representing semantic networks for the distribution and use amongst computers \cite{rdfintro:miller1998,rdfcon:klyne2004,rdfspec:manola2004}. The Resource Description Framework Schema (RDFS) is a popular ontology language for RDF \cite{rdfs:brickley2004}. An RDF network can be represented as a triple list $G^n \subseteq (V \times \Omega \times V)$, where $\Omega$ is a set of edge labels denoting the semantic (or meaning) of the relationship between the vertices in $V$ and any ordered triple $\la i, \omega, j \ra \in G^n$ states that vertex $i$ is related to vertex $j$ by the semantic $\omega$. The use of labeled edges complicates the meaning of the rank vector returned by single-relational centrality measures because some vertices may be deemed more central than others with respect to one edge label, but not with respect to another. For example, the relationship \texttt{isFriendOf} may be considered more relevant than \texttt{livesInSameCityAs}. Therefore, due to the number of ways by which two adjacent vertices can be related and the focus on the semantics of such relations, the aim of recent semantic network metrics have been on ranking semantic associations \cite{semantic:rada1989,discov:lin2004,semassoc:sheth2005,semrank:boan2005}, not the vertices themselves. A semantic association between vertices $i$ and $j$ is defined by the ordered multi-set path $q$, where $q = (i, \omega_a, \ldots, \omega_b, j)$, $i,j \in V$, and $\omega_a,\omega_b \in \Omega$ \cite{rhoquery:anyanwu2003}. If $Q_{i,j}$ is the set of all possible semantic associations between vertices $i$ and $j$ in $G^n$, then a path metric function is generally defined as $f : Q_{V,V} \rightarrow \mb{R}^{|Q_{V,V}|}$, where the range of $f$ denotes the ranking of each path in $Q_{i,j}$.

This article focuses on vertex ranking, not path ranking. Moreover, this article is primarily interested in eigenvector-based metrics such as eigenvector centrality \cite{power:bonacich1987} and PageRank \cite{page98pagerank}. While eigenvector-based metrics on semantic networks have been proposed to rank vertices, the algorithms rely on prescribed semantic network ontologies and therefore, have not been generalized to handle any semantic network instance \cite{ranksem:zhuge2003,pagesem:mihalcea2004,socialgrammar:rodriguez2007}. This article presents a method for applying eigenvector-based centrality metrics to semantic networks such that the semantic network's ontology is respected. The proposed method extends the random walker model of Markov chain analysis \cite{markov:haggstrom2002} to support its application to semantic network vertex ranking without altering the original data set or isolating subsets of the data set for analysis. This method is called the grammar-based random walker method. While the random walker's of Markov chain analysis are memoryless, grammar-based random walkers of semantic networks utilize a user-defined grammar (or program) that instructs the grammar-based random walker to take particular ontological paths through the semantic network instance. Moreover, a grammar-based random walker maintains a memory of its path in the network and in the grammar in order for it to execute simple logic along its path. This simple logic allows the grammar-based random walker to generate semantically complex eigenvector rankings. For example, given a scholarly semantic network and the grammar-based method, it is possible to calculate $\vpi$ over all author vertices such that the authors indexed by $\vpi$ are located at some institution and they wrote an article that cites another article of a different author of the same institution.

The next section provides an overview of the class of eigenvector-based metrics for single-relational networks that use the random walker model and then proposes a method for meaningfully applying such metrics to semantic networks. The result is a vertex valuing function generally defined as $f : G \times \Psi \rightarrow  \mathbb{R}^{|\subseteq V|}$ where $\Psi$ is a user defined grammar and $\vpi \in \mathbb{R}^{|\subseteq V|}$.

\section{Random Walkers in Single-Relational Networks}

The random walker model comes from the field of Markov chain analysis. Markov chains are used to model the dynamics of a stochastic system by explicitly representing the states of the system and the probability of transition between those states \cite{mitrani:prob1998,markov:ching2006}. A Markov chain can be represented by a directed weighted network $G^1 = (V,E , w)$ where the set of vertices in $V$ are system states, $E \subseteq V \times V$ are the set of directed edges representing the transitions between states, and $w: E \rightarrow [0,1]$ is the function that maps each edge to a real weight value that represents the state transition probability \footnote{Note that while the weight function $w$ does in fact label edges in $E$, the meaning of the edges are homogenous and thus, $\omega$ simply denotes the extent to which the meaning is applied. Therefore, with respects to this article, a weighted Markov chain is considered a single-relational network, not a semantic network.}. The outgoing edge weights of any state in the Markov chain form a probability distribution such that $\sum_{e \in \Gamma^+(i)} w(i) = 1 \; : \; |\Gamma^+(i)| \geq 1$, where $\Gamma^+(i) \subseteq E$ is the set of outgoing edges of vertex $i$. The future state of the system at time $n+1$ is based solely on the current state of the system at time $n$ and its respective outgoing edges.

Given that a Markov chain can be represented by a weighted directed network, one can envision a random walker moving from vertex to vertex (i.e.~state to state). A random walker moves through the Markov chain by choosing a new vertex according to the transition probabilities outgoing from its current vertex. This process continues indefinitely where the long run behavior, or stationary distribution denoted $\vpi$, of the random walker makes explicit the probability of the random walker being located at any one vertex at some random time in the future. However, only aperiodic, irreducible, and recurrent Markov chains can be used to generate a $\vpi$ that is the stationary distribution of the chain \cite{netanal:brandes2005}. If the Markov chain is aperiodic then the random walker does not return to some previous vertex in a periodic manner. A Markov chain is considered recurrent and irreducible if there exists a path from any vertex to any other vertex. In the language of graph theory, the weighted directed network representing the Markov chain must be strongly connected. If $\mathbf{A} \in \mathbb{R}^{|V| \times |V|}$ is the weighted adjacency matrix representation of $G^1$ and there exists a vertex vector $\vpi \in \mathbb{R}^{|V|}$ where $\sum_{i \in V} \vpi_i = 1$ and $\mathbf{A}\vpi = \lambda\vpi$, where $\lambda$ is the greatest eigenvalue of all eigenvectors of $\mathbf{A}$, then $\vpi$ is the stationary distribution of $G^1$ as well as the primary eigenvector of $\mathbf{A}$ \cite{parzen:stochastic1962}. The vector $\vpi$ represents the eigenvector centrality values for all vertices in $V$ \cite{power:bonacich1987}.

In the real world, periodicity is highly unlikely in most natural networks \cite{netanal:brandes2005}. However, a strongly connected network is not always guaranteed. If the network is not strongly connected, then the problem of rank sinks and subset cycles is introduced and $\vpi$ is not a real valued vector.  Therefore, many networks require some manipulation to ensure strong connectivity. For example, the web citation network, represented as $G^1 = (V,E)$, is not strongly connected \cite{bowtie:broder} and therefore, in order to calculate $\vpi$ for the web citation network, it is necessary to transform $G^1$ into a strongly connected network. One such method was introduced in \cite{anatom:brin1998,page98pagerank} where a probabilistic web citation network is overlaid with a fully connected web citation network. In matrix form, the probabilistic adjacency matrix of the web citation network, $\mathbf{A} \in \mathbb{R}^{|V| \times |V|}$, is created, where
%%%
\begin{equation*}
\mathbf{A}_{i,j} =
	\begin{cases}
		\frac{1}{|\Gamma^+(i)|} & \text{if } (i,j) \in E \\
		\frac{1}{|V|} & \text{if } |\Gamma^+(i)| = 0.
	\end{cases}
\end{equation*}
%%%
In $\mathbf{A}$, all rank sinks (i.e.~vertices with no out degree, absorbing vertices) connect to every other vertex in $V$ with equal probability. Next, the matrix $\mathbf{B}$ is created such that $\mathbf{B} \in \mathbb{R}^{|V| \times |V|}$ and $\mathbf{B}_{i,j} = \frac{1}{|V|}$ for all $i$ and $j$ in $V$. $\mathbf{B}$ denotes a fully connected network (i.e.~a complete network) where every vertex is connected to every other vertex with equal probability. The composite adjacency matrix $\mathbf{C} = \delta \mathbf{A} + (1-\delta) \mathbf{B}$, where $\delta \in (0,1]$ is a parameter weighting the contribution of each adjacency matrix, guarantees that there is some finite probability that each vertex in $V$ is reachable by every other vertex in $V$. Therefore, the network denoted by $\mathbf{C}$ is strongly connected and there exists a unique stationary distribution $\vpi$ such that $\mathbf{C} \vpi = \lambda\vpi$. This method of inducing strong connectivity is called PageRank and has been used extensively to rank vertices in a unlabeled, single-relational networks \cite{google:langville2006}.

The primary contribution of this article is that it ports the eigenvector-based algorithms of single-relational networks over to the semantic network domain. This article presents a method for calculating a semantically meaningful stationary distribution within some subset of a semantic network (called grammar-based eigenvector centrality) as well as how to implicitly induce strong connectivity irrespective of the network's topology (called grammar-based PageRank). This general method is called the grammar-based random walker model because a random walker does not blindly move from vertex to vertex, but instead is constrained by a grammar that ensures that the stationary distribution is calculated in a ``grammatically correct" subset of $G^n$. Before discussing the grammar-based random walker method, the next section provides a brief review of semantic networks, ontologies, and current standards for their representation.

\section{Semantic Networks}

A semantic network is also known as a multi-relational network or directed labeled network. In a semantic network, there exists a heterogeneous set of vertex types and a heterogeneous set of edge types such that any two vertices in the network can be connected by zero or more edges. In order to make a distinction between two edges connecting the same vertices, a label denotes the meaning, or semantic, of the relationship. A semantic network can be represented by the triple list $G^n \subseteq (V \times \Omega \times V)$. A vertex to vertex relationship is called a triple because there exists the relationship $\la i, \omega, j \ra$ where $i \in V$ is called the subject, $\omega \in \Omega$ is called the predicate, and $j \in V$ is called the object.

Perhaps the most popular standard for representing semantic networks is the Resource Description Framework (RDF) of the Semantic Web initiative \cite{rdfspec:manola2004,rdfcon:klyne2004}. There currently exists many applications to support the creation, query, and manipulation of RDF-based semantic networks. High-end, modern day triple-stores (RDF databases) can reasonably support on the order of $10^9$ triples \cite{agraph:aasman2006}. For this reason, and due to the fact that RDF is becoming a common data model for various disciplines including digital libraries \cite{lib:bax2004}, bioinformatics \cite{sembio:ruttenberg2007}, and computer science \cite{semturing:rodriguez2007}, all of the constructs of the grammar-based random walker model will be presented according RDF and its ontology modeling language RDFS.

\begin{figure}[h!]
	\centering
	\includegraphics[width=0.35\textwidth]{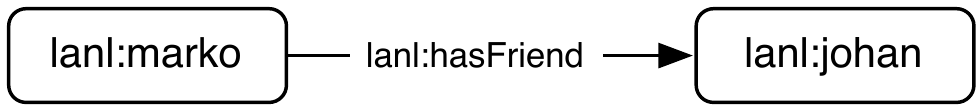}
	 \caption{\label{fig:friend-example}A example triple in RDF.}
\end{figure}

RDF identifies vertices in a semantic network by Uniform Resource Identifiers (URI) \cite{uri:berners2005}, literals, or blank nodes (also called anonymous nodes) and edge labels are represented by URIs. An example RDF triple where all components are URIs is
%%%
\begin{equation*}
	\la \ttt{lanl:marko}, \ttt{lanl:hasFriend}, \ttt{lanl:johan} \ra.
\end{equation*}
%%%
In this triple, \ttt{lanl} is a namespace prefix that represents \ttt{http://www.lanl.gov}. This prefix convention is used throughout the article to ensure brevity of text and diagram clarity. Figure \ref{fig:friend-example} is a graphic representation of the previous triple.
%%%
 
Another example of a triple where the object is a literal is
%%%
\begin{equation*}
	\la \ttt{lanl:marko}, \ttt{lanl:hasFirstName}, \ttt{"Marko"$^\wedge$$^\wedge$xsd:string} \ra.
\end{equation*}
%%%
In this triple, the literal \ttt{"Marko"$^\wedge$$^\wedge$xsd:string} is an XML schema datatype string (\ttt{xsd}) \cite{xsd:biron2004}.

While a semantic network instance is represented in pure RDF, a semantic network ontology is represented in RDFS (a language represented in RDF).

\subsection{Ontologies}

Due the heterogeneous nature of the vertices and edges in a semantic network, an ontology is usually defined as way of specifying the range of possible interactions between the vertices in the network. Ontologies articulate the relation between abstract concepts and make no explicit reference to the instances of those classes \cite{semdef:sowa1987}. For example, the ontology for the web citation network can be defined by a single class representing the abstract concept of a web page and the single semantic relationship representing a web link or citation (i.e.~ \texttt{href}). This simple ontology states that the network representing the semantic model of the web is constrained to only instances of one class (a web page) and one relationship (a web link). 

\begin{figure}[h!]
	\centering
	\includegraphics[width=0.325\textwidth]{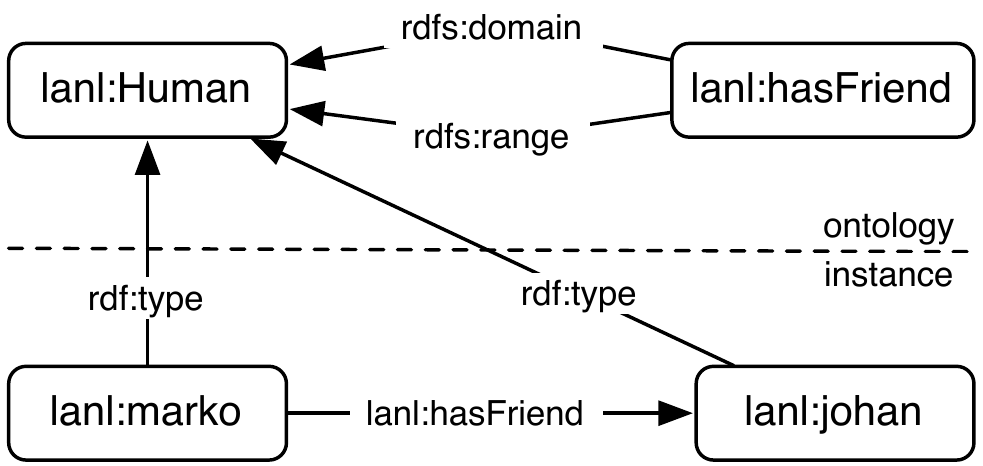}
	 \caption{\label{fig:friend-ontology}A example of the relationship between an ontology and its instance.}
\end{figure}

Given the previous single triple represented in Figure \ref{fig:friend-example}, the semantic network ontology could be represented as diagramed in Figure \ref{fig:friend-ontology}, where the \ttt{lanl:hasFriend} property must have a domain of \ttt{lanl:Human} and a range of \ttt{lanl:Human}, where \ttt{lanl:marko} and \ttt{lanl:johan} are both \ttt{lanl:Human}s.
%%%

\begin{figure}[h!]
	\centering
	\includegraphics[width=0.325\textwidth]{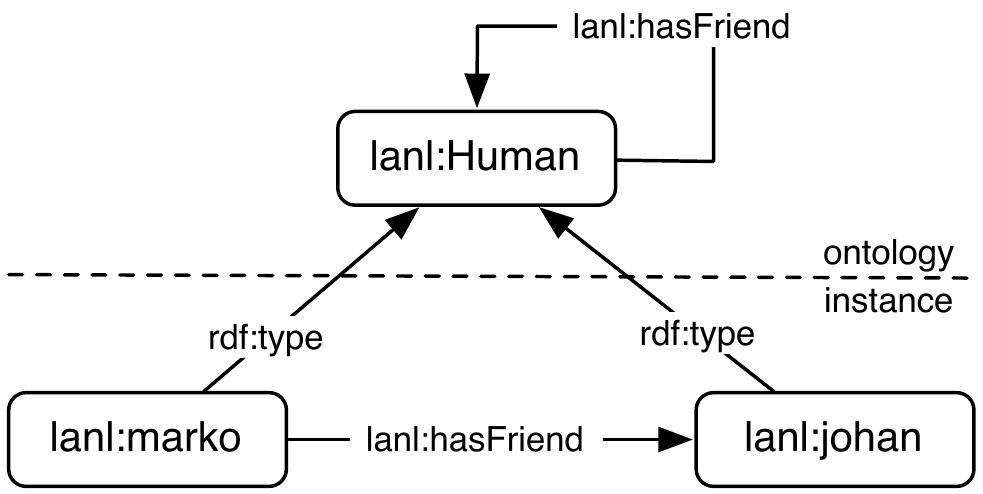}
	 \caption{\label{fig:friend-abbrev}An abbreviation of the diagramed in Figure \ref{fig:friend-ontology}.}
\end{figure}

Note that ontological diagrams can be abbreviated by assuming that the tail of an edge is the \ttt{rdfs:domain} and the head of the edge is the \ttt{rdfs:range}. This abbreviated form is diagrammed in Figure \ref{fig:friend-abbrev}.
%%%

\begin{figure}[h!]
	\centering
	\includegraphics[width=0.3\textwidth]{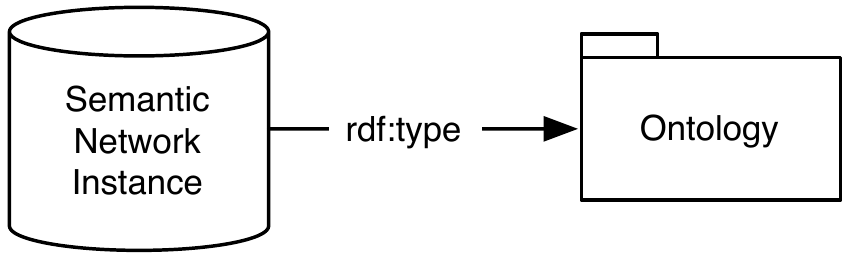}
	 \caption{\label{fig:ont-inst}The relationship between a semantic network instance and its ontology.}
\end{figure}

In general, the relationship between an ontology and its corresponding semantic network instantiation is depicted in Figure \ref{fig:ont-inst} where the \texttt{rdf:type} property denotes that the vertices in $V$ are an instance of some abstract class in the ontology.

RDFS does not provide a large enough vocabulary to describe many of the types of relations needed for modeling class interactions \cite{owl:lacy2005}. For this reason, other modeling languages, based on RDFS, have been developed such as the Web Ontology Language (OWL) \cite{owlspec:mcguinness2004,owl:lacy2005}. OWL allows a modeler to represent restrictions on properties (e.g.~cardinality) and provides a broader range of property types (e.g.~inverse relationships, functional relationships). Even though RDFS is limited in its expressiveness it will be used as the modeling language for describing the grammar-based random walker ontology. Note that it is trivial to map the presented concepts over to other modeling languages such as OWL. For a more in-depth review of ontology modeling languages, their history, and their application, please refer to \cite{owl:lacy2005} and \cite{ont:gasevic2006}.

The next section brings together the concepts of random walkers, semantic networks, and ontologies in order to formalize this article's proposed grammar-based random walker model.

\section{Grammar-Based Random Walkers}
 
A grammar-based random walker moves through a semantic network in a manner that respects the labels of the edges  connecting the network's vertices. The purpose of the grammar-based random walker is to identify the stationary distribution of some subset of the full semantic network (i.e.~the primary eigenvector of a sub-network of the network). Unlike the random walkers of Markov chain analysis, a grammar-based random walker does not take any outgoing edge from its current vertex, but instead, depending on the user defined grammar, traverses particular edges types to particular vertex types.

\begin{figure}[h!]
	\centering
	\includegraphics[width=0.45\textwidth]{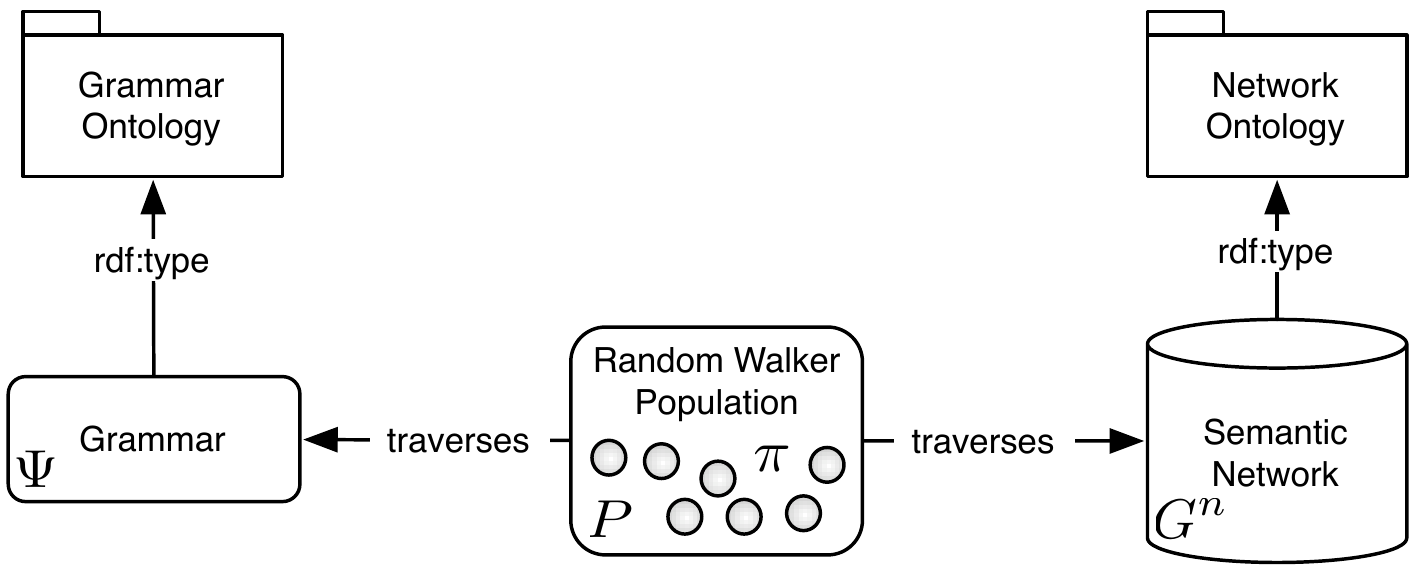}
	 \caption{\label{fig:sys-arch}The grammar-based random walker architecture.}
\end{figure}

Any designed grammar uses the constructs and algorithms defined by the grammar ontology (prefixed as \texttt{rwr}). The grammar ontology defines rule classes, attribute classes, data structures, and properties that are intended to be combined with instances and classes of $G^n$ to create a $G^n$ specific grammar denoted $\Psi$. The rules of the grammar ultimately determine which vertices in $V$ are indexed by the returned rank vector $\vpi$. The rank vector $\vpi$ is created by a set of grammar-based random walkers $P$ traversing through $G^n$ and obeying $\Psi$. Figure \ref{fig:sys-arch} diagrams the relationship between $\Psi$, $P$, $G^n$, and their respective ontologies. Note that $\Psi$, $\Psi$'s ontology, $G^n$, and $G^n$'s ontology are all semantic networks and thus, can be represented by the same semantic network data structure. However, in order to make the separation between the components clear, each data structure will be discussed as a separate semantic network.
%%%

The meaning of the vertex rank vector $\vpi$ of the grammar-based model, both semantically and theoretically, depends primarily on the grammar used. Some $\Psi$s will generate a $\vpi$ that is the stationary probability distribution of some subset of $G^n$, while others will be more representative of a discrete form of the spreading activation models, where calculating the long run behavior of the random walker is undesirable \cite{inform:cohen1987,spread:savoy1992,applic:crestani1997,search:crestani2000}. In practice, determining whether $\vpi$ is a stationary distribution of the analyzed subset of $V$ is a matter of determining whether the subset of $G^n$ that is traversed by $P$ is strongly connected and the normalized $\vpi$ has converged to a stable set of values. Any grammar-based random walker implementation is a function generally defined as $f : G \times \Psi \rightarrow \mathbb{R}^{|\subseteq V|}$.

It is noted that there exists two related ontologies for modeling the distribution of discrete entities in a semantic network. These ontologies were inspirational to the ideas presented in this article. The marker passing Petri net ontology of \cite{petri:gasevic2006} and the particle swarm ontology of \cite{socialgrammar:rodriguez2007}. However, both ontologies were designed for a different application space. The first is for Petri net algorithms while the latter was defined specifically for collective decision making systems. Finally, the grammar-based model presented in \cite{geodesics:rodriguez2007} for calculating geodesics in a semantic network combined with the grammar-based model presented in this article form a unified framework for porting many of the popular single-relational network analysis algorithms over to the semantic network domain (more specifically, the RDF and Semantic Web domain).

\subsection{The Grammar-Based Random Walker Ontology}

\begin{figure}[h!]
	\centering
	\includegraphics[width=0.49\textwidth]{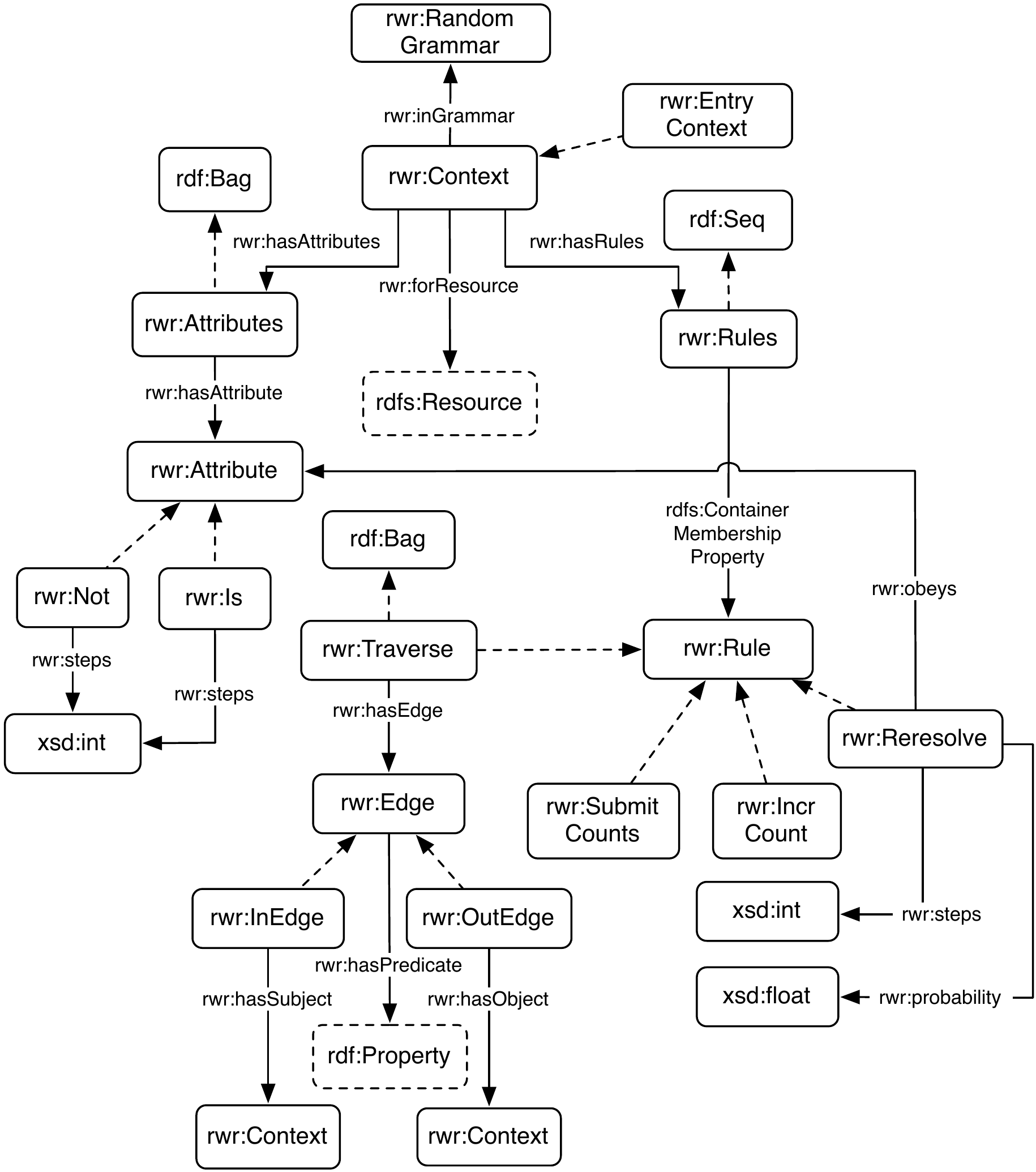}
	 \caption{\label{fig:ont-rwg}The complete grammar-based random walker ontology.}
\end{figure}

The complete grammar ontology is graphically represented in Figure \ref{fig:ont-rwg}, where squares are \texttt{rdfs:Class}es and edge labels are \texttt{rdf:Property} types. The tail of each edge is the \ttt{rdfs:domain} of the \ttt{rdf:Property} and the head is the \ttt{rdfs:range}. For the purpose of diagram clarity, the dashed edges denote a relationship of \texttt{rdfs:subClassOf}. Finally, note that the two dashed squares should be instances or classes that are in $G^n$ or its ontology, respectively.
%%%

The grammar ontology follows a convention similar to most object oriented programming languages \cite{proglang:sebesta2005} in that a \texttt{rwr:Context} (i.e.~class) has a set of attributes (i.e.~fields) and rules (i.e.~methods). The general idea is that any grammar instance $\Psi$ is a collection of \texttt{rwr:Context} objects connected to one another by \texttt{rwr:Traverse} rules. \texttt{rwr:Context}s and their \texttt{rwr:Traverse} rules are an abstract model of what triples a grammar-based random walker can traverse in $G^n$. The \texttt{rwr:Is} and \texttt{rwr:Not} attributes further constrain the types of vertices that can be traversed by the  random walker and are used for path ``bookkeeping" and path logic. The \texttt{rwr:IncrCount} and \texttt{rwr:SubmitCounts} rules determine which vertices in $V$ should be indexed by $\vpi$. Finally, the \texttt{rwr:Reresolve} rule is the means by which the random walker is able to ``teleport" to other regions of $G^n$. The \texttt{rwr:Reresolve} rule is used to model the PageRank algorithm and therefore, is a mechanism for guaranteeing that the subset of $G^n$ that is traversed is strongly connected and $\vpi$ is a stationary distribution.

\subsection{High-Level Overview of the Grammar-Based Model}

This section will provide a high-level overview of the components of the grammar diagrammed in Figure \ref{fig:ont-rwg}. $\Psi$ is a user defined data structure that is created specifically for $G^n$ and $G^n$'s respective ontology. Any $\Psi$ must obey the constraints defined by the grammar ontology diagrammed in Figure \ref{fig:ont-rwg}. A single grammar-based random walker (denoted $p \in P$) ``walks" both $G^n$ and $\Psi$ in order to  dynamically generate a vertex rank vector denoted $\vpi$. If the $p$-traversed subset of $G^n$ is strongly connected, then only a single random walker is needed to compute $\vpi$ \cite{markov:haggstrom2002}.

When random walker $p \in P$ is at some \texttt{rwr:Context} in $\Psi$, the \texttt{rwr:Context} is ``resolved" to a particular vertex in $V$. This is the relationship between $\Psi$ and $G^n$. For example, if $p$ is at some \ttt{rwr:Context} in $\Psi$ that is \ttt{rwr:forResource} \ttt{lanl:Human}, then $p$ must also be at some vertex in $V$ that is of \ttt{rdf:type} \ttt{lanl:Human}. Thus, $\Psi$ is an abstract representation of the legal vertices that $p$ can traverse in $V$. When $p$ is at a \ttt{rwr:Context}, $p$ will execute the \texttt{rwr:Context}'s collection of \texttt{rwr:Rule}s, while at the same time respecting \ttt{rwr:Context} \ttt{rwr:Attribute}s. The collection of \ttt{rwr:Rule}s is an ordered \texttt{rdf:Seq} \cite{rdfs:brickley2004}. This means that $p$ must execute the rules in their specified sequence. This is represented as the set of properties \texttt{rdf:\_1}, \texttt{rdf:\_2}, \texttt{rdf:\_3}, etc. (i.e.~\texttt{rdfs:subPropertyOf} \texttt{rdfs:ContainerMembershipProperty}).

Any grammar-based random walker $p$ has three local variables: 
%%%
\begin{itemize}
	\item a reference to its path history in $G^n$ (denoted $g^p$)
	\item a reference to its path history in $\Psi$ (denoted $\psi^p$)
	\item a local vertex vector (denoted $\vpi^p \in \mathbb{N}^{|\subseteq V|}$)
\end{itemize}
%%%
and a reference to a single global variable:
%%%
\begin{itemize}
	\item a global vertex vector (denoted $\vpi \in \mathbb{N}^{|\subseteq V|}$)
\end{itemize}

The path history $g^p$ is an ordered multi-set of vertices, edge labels, and edge directionalities. If the random walker $p$ traversed the path diagrammed in Figure \ref{fig:friend-example} from left to right, then $g^p = \{ \ttt{lanl:marko}, \ttt{lanl:hasFriend}, +, \ttt{lanl:johan} \}$. Note that $g^p_0 = \ttt{lanl:marko}$, $g^p_{1'} = \ttt{lanl:hasFriend}$, $g^p_{1''} = +$, and $g^p_{1} = \ttt{lanl:johan}$, where $n'$ denotes the edge label used to get to the vertex at time $n$ and $n''$ denotes the direction that $p$ traversed over that edge. In the grammar-based random walker model, a random walker can, if stated in $\Psi$, oppose an edge's directionality. For example, if $p$ had traversed the edge diagrammed in Figure \ref{fig:friend-example} from right to left, then $g^p = \{ \ttt{lanl:johan}, \ttt{lanl:hasFriend}, -, \ttt{lanl:marko} \}$. A similar convention holds for $p$'s $\Psi$-history $\psi^p$. However, in $\psi^p$ the vertices are \ttt{rwr:Context}s, the edge labels are the \ttt{rdf:Property} of the \ttt{rwr:Edge} chosen, and the directionalities are determined by whether an \ttt{rwr:OutEdge} or \ttt{rwr:InEdge} was traversed.

The ``walking" aspect of $p$ for both $\Psi$ and $G^n$ is governed by the \texttt{rwr:Traverse} rule. When $p$ executes a \texttt{rwr:Traverse} rule in $\Psi$, it selects a particular \texttt{rwr:Edge} to traverse. For \texttt{rwr:OutEdge}s, a triple in $G^n$ is selected with the subject being its current location $g^p_n$, and predicate and objects are instances of the respective resource specified by the \texttt{rwr:OutEdge} (\ttt{rwr:hasPredicate} and \ttt{rwr:hasObject}). For \texttt{rwr:InEdge}s, a triple in $G^n$ is selected where $g^p_n$ is the object of the triple and the subject and predicate are instances of the resource specified by the \texttt{rwr:InEdge} (\ttt{rwr:hasPredicate} and \ttt{rwr:hasSubject}). The \texttt{rwr:Context} chosen is $\psi^p_{n+1}$ and the \texttt{rdfs:Resource} of the triple $\langle \psi^p_{n+1}, \texttt{rwr:forResource}, ?x \rangle \in \Psi$ determines $g^p_{n+1}$, where $?x$ is any class in $G^n$'s ontology or instance in $G^n$'s vertex set $V$. The newly chosen $g^p_{n+1}$ is called the resolution of $\psi^p_{n+1}$.

The \texttt{rwr:IncrCount} and \texttt{rwr:SubmitCounts} rules effect the random walker's local vertex vector $\vpi^p$ and the  global vertex vector $\vpi$, respectively. The distinction between $\vpi^p$ and $\vpi$ is that $\vpi^p$ is a temporary counter that is not submitted to the global counter $\vpi$ until the \ttt{rwr:SubmitCounts} rule has been executed. The walker $p$ does not submit its vertex counts until it has determined that it is in a $\Psi$-correct subset of $G^n$.

The process of moving $p$ through a semantic network and allowing it to increment a counter for specific vertices continues until the ratio between the values of the global $\vpi$ converge. Note that $\vpi$ does not provide a probability distribution, $\sum_{i \in \vpi} \vpi_i \neq 1$. Instead, $\vpi$ represents the number of times an indexed vertex of $\vpi$ has been counted  by a grammar-based random walker. Therefore, to determine the probability of being at any one vertex that is indexed by $\vpi$, $\vpi$ can be normalized to generate a new vector denoted $\vpi' \in \mathbb{R}^{|\subseteq V|}$, where $\vpi'_i = \frac{\vpi_i}{\sum_{j \in \vpi} \vpi_j}$. If $\vpi'$ is the normalization of $\vpi$ then, when $\vpi'$ no longer changes with successive executions of the \texttt{rwr:SubmitCounts} rule, the process is complete. More formally, if $\epsilon \in \mathbb{R}$ is an argument specifying the smallest change accepted for convergence consideration, then the grammar-based random walker algorithm is complete when ${|| \vpi'_{_n} -  \vpi'_{_m} ||}_2 < \epsilon$, where $n$ and $m$ are the time steps of consecutive calls to \texttt{rwr:SubmitCounts}. However, like Markov chains, this convergence will only occur if the subset of $G^n$ that is traversed is strongly connected and aperiodic. If the traversed subset of $G^n$ is not strongly connected or is periodic, then the \texttt{rwr:Reresolve} rule can be used to simulate grammar-based random walker ``teleportation". With the inclusion of the \ttt{rwr:Reresolve} rule, a grammar-based PageRank can be executed on $G^n$. 

The next section will formalized each of the \ttt{rwr:Rule}s and \ttt{rwr:Attribute}s of the grammar ontology.

\section{The Rules and Attributes of the Grammar Ontology}

The following \ttt{rwr:Rule}s and \ttt{rwr:Attribute}s are presented in a set theoretic form that borrows much of its structure from semantic query languages such as SPARQL \cite{sparql:prud2004}. The query triple $\langle ?x, \texttt{rdf:type}, \texttt{lanl:Author} \rangle \in G^n$ will bind $?x$ to any \texttt{lanl:Author} in the semantic network $G^n$. The $?x$ notation represents that $?x$ is a variable that is bound to any vertex (i.e.~URI) that matches the query pattern. The same query can return many resources that bind to $?x$. In such cases, the results are returned as a set. Thus $X = \{?x \; | \; \langle ?x, \texttt{rdf:type}, \texttt{lanl:Author} \rangle \in G^n \}$ denotes the set of all vertices in $V$ that are of \ttt{rdf:type} \texttt{lanl:Author}.

The following subsections present each of the \ttt{rwr:Rule}s and \ttt{rwr:Attribute}s that a grammar-based random walker must execute and respect during its journey through both $\Psi$ and $G^n$.

\subsection{Entering $\Psi$ and $G^n$}

Every random walker ``walks" both $\Psi$ and $G^n$ in parallel. However, before a walker can walk either data structure, it must enter both $\Psi$ and $G^n$. The entry points of $\Psi$ are \texttt{rwr:EntryContext}s and are represented by the set $s(\Phi)$, where
%%%
\begin{align*}
	s(\Phi) =& \{ ?x \; | \; \langle ?x, \texttt{rdf:type}, \texttt{rwr:EntryContext} \rangle \in \Psi \}.
\end{align*}
%%%
The starting location $\phi \in s(\Phi)$ of $p$ is chosen with probability $\frac{1}{|s(\Phi)|}$. Once some $\phi$ is chosen, $\psi^p_0=\phi$ (time $n$ starts at $0$). An entry location into $V$ can be determined by randomly selecting some vertex $i \in s(V \; | \; \phi)$, where $s(V \; | \; \phi)$ is the set of all $i \in V$ given that $i$ is a proper resolution of the \texttt{rwr:EntryContext} $\phi$. Thus,
%%%
\begin{align*}
	 s(V \; | \; \phi) =& \{ ?i \; | \; \langle \phi, \texttt{rwr:forResource}, ?z \rangle \in \Psi \\
			         & \; \wedge \; ( \langle ?i, \texttt{rdf:type}, ?z \rangle \in \Psi \; \vee \;  ?i = ?z)  \},
\end{align*}
%%%
where type inheritance is strictly followed. For instance, if $i$ is an \texttt{rdf:type} of $z$ then $i$ is an instance of $z$ or an instance of $u$ where $u$ is a \texttt{rdfs:subClassOf} $z$. This is subsumption in RDFS reasoning and will be used repeatedly throughout the remainder of this article.

Given the set $s(V \; | \; \phi)$, the probability of $p$ choosing some $ i \in s(V \; | \; \phi)$ is $\frac{1}{|s(V \; | \; \phi)|}$. The chosen vertex $i$ becomes the starting location of $p$ in $G^n$ and thus, $g^p_{0} = i$.

\begin{figure}[h!]
	\centering
	\includegraphics[width=0.25\textwidth]{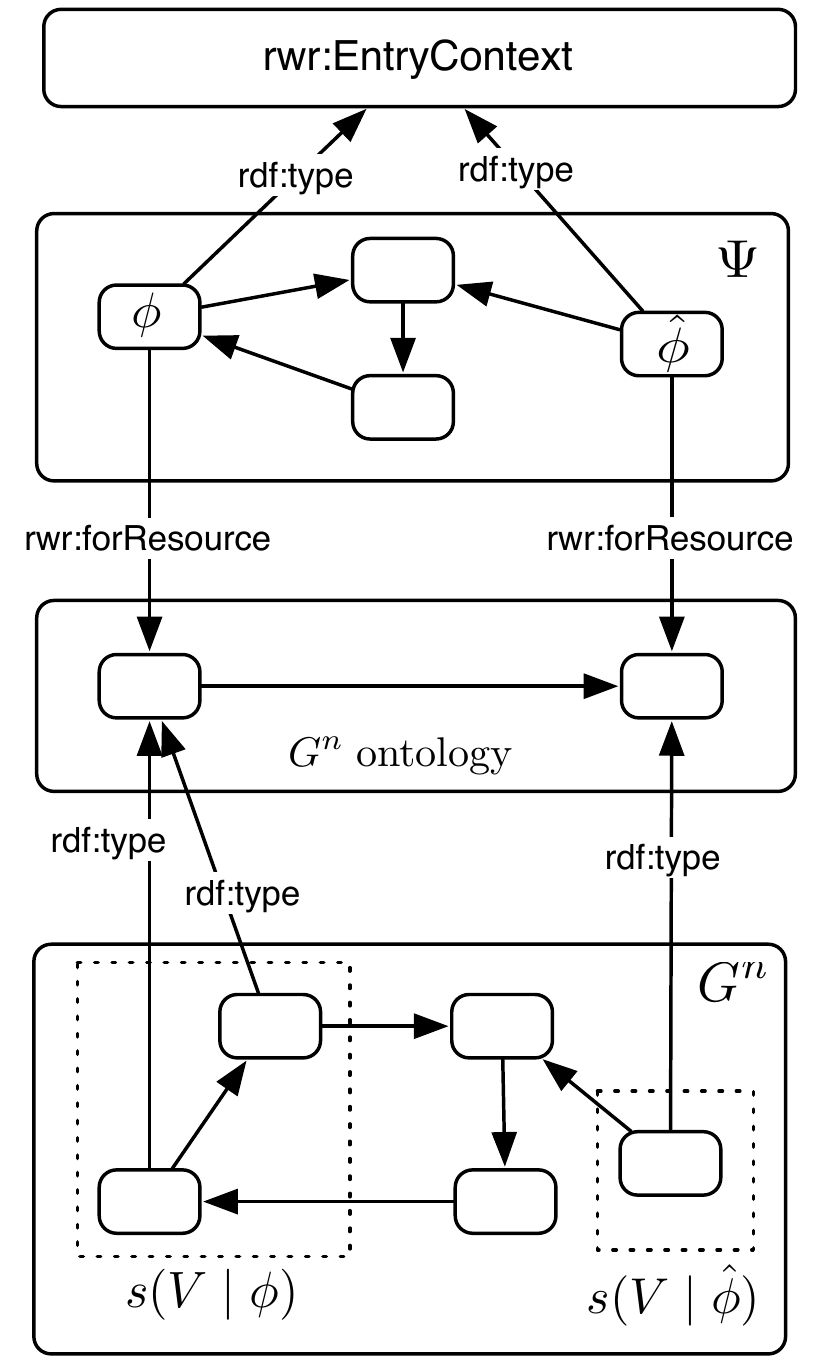}
	 \caption{\label{fig:entry-example}The relationship between \texttt{rwr:EntryContext}s in $\Psi$, $G^n$, and $G^n$'s ontology.}
\end{figure}

Note that $g^p_{0'} = \emptyset$, $\psi^p_{0'} = \emptyset$, $g^p_{0''} = \emptyset$ and $\psi^p_{0''} = \emptyset$ since a random walker enters both $\Psi$ and $G^n$ at a vertex without using an intervening edge label or directionality. Figure \ref{fig:entry-example} depicts how \texttt{rwr:EntryContext}s in $\Psi$ are related to vertices in $G^n$.
%%%

\subsection{The \ttt{rwr:Not} Attribute}

Before presenting the \ttt{rwr:Traverse} rule, it is important to discuss the two attributes that constrain the \ttt{rwr:Traverse} rule: namely, \ttt{rwr:Not} and \ttt{rwr:Is}. This subsection will discuss the \ttt{rwr:Not} attribute. The next section will discuss the \ttt{rwr:Is} attribute. The \ttt{rwr:Not} atttribute ensures that the random walker $p$ does not traverse an edge to a particular, previously seen vertex in $g^p$. Any \ttt{rwr:Not} attribute is the subject of a triple with a predicate \ttt{rwr:steps} and literal $m \in \mb{N}$. The literal $m$ denotes which vertex from $m$-steps ago $p$ must avoid. In other words, $p$ must not have a $g^p_{n+1}$ that equals $g^p_{n-m}$. Thus, the \texttt{rwr:Context} $\psi^p_{n+1}$ cannot resolve to $g^p_{n-m}$. If
%%%
\begin{align*}
M = & \{?m \; | \; \langle \psi^p_{n+1}, \texttt{rwr:hasAttributes}, ?x \rangle \in \Psi \\
       & \; \wedge \; \langle ?x, \texttt{rwr:hasAttribute}, ?y \rangle \in \Psi \\
       & \; \wedge \; \langle ?y, \texttt{rdf:type}, \texttt{rwr:Not} \rangle \in \Psi \\
       & \; \wedge \; \langle ?y, \texttt{rwr:steps}, ?m \rangle \in \Psi \},
\end{align*}
%%%
then
%%%
\begin{align*}
X(p)_{n+1} = \bigcup_{m \in M} g^p_{n-m},
\end{align*}
%%%
where $X(p)_{n+1} \subseteq V$ and $X(p)_{n+1} \cap g^p_{n+1} = \emptyset$. The set $X(p)_{n+1}$ is the set of vertices in $V$ that $g^p_{n+1}$ must not equal.

\begin{figure}[h!]
	\centering
	\includegraphics[width=0.275\textwidth]{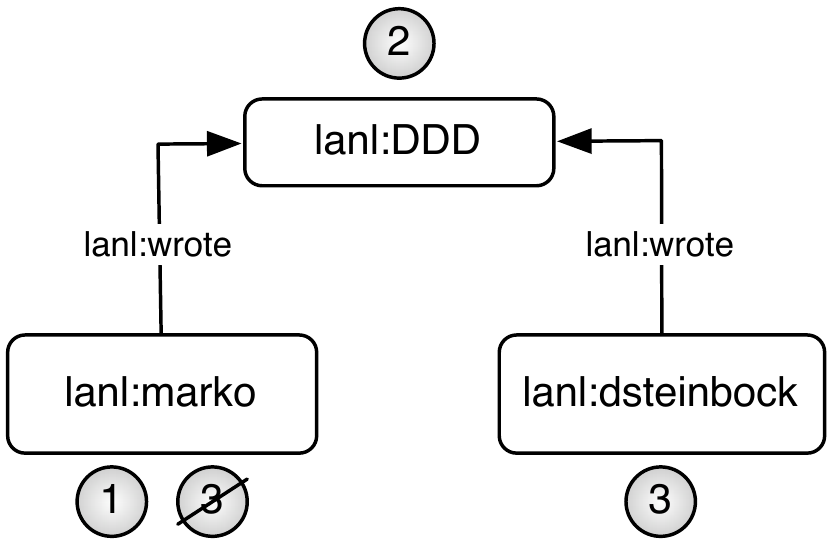}
	 \caption{An example situation for the \texttt{rwr:Not} attribute \label{fig:not-example}}
\end{figure}

The \texttt{rwr:Not} attribute is useful when $p$ must not return to a vertex in $V$ that has been previously visited. Imagine that $p$ is determining whether or not a particular article has at least two authors (or must traverse an implicit coauthorship network). Such an example is depicted in Figure \ref{fig:not-example}, where the numbered circles are the location of $p$ at particular time steps and author vertices are only connected to their authored articles. If, at $n=1$, $p$ is located at \texttt{lanl:marko} then $p$ will traverse the \texttt{lanl:wrote} predicate to the \texttt{lanl:DDD} article. If $p$ is checking for another author that is not \texttt{lanl:marko} then $p$ can only take the \texttt{lanl:wrote} predicate to \texttt{lanl:dsteinbock}. If \texttt{lanl:DDD} only had one author, then $p$ would be stuck (i.e.~halt) at \texttt{lanl:DDD} since no legal \texttt{lanl:wrote} predicate could be traversed. At which point, it is apparent that the article has only one author. Moreover, by traversing to \ttt{lanl:dstreinbock} and not back to \ttt{lanl:marko} at $n=3$, a coauthorship network is implicitly traversed.
%%%

\subsection{The \ttt{rwr:Is} Attribute}

Unlike the \ttt{rwr:Not} attribute, the \ttt{rwr:Is} atttribute is used to ensure that the random walker $p$ does, in fact, traverse an edge to a previously visited vertex in $V$. Any \ttt{rwr:Is} attribute is the subject of a triple with a predicate \ttt{rwr:steps} and literal $m \in \mb{N}$. The literal $m$ denotes which vertex from $m$-steps ago $p$ must traverse to. If this set of vertices returned by the \ttt{rwr:Is} attribute is greater than $1$, then $p$ must traverses to one of the vertices from the set. Thus, the random walker $p$ must have vertex $g^p_{n+1}$ equal some $g^p_{n-m}$. In other words, the \texttt{rwr:Context} $\psi^p_{n+1}$ must resolve to some $g^p_{n-m}$. If
%%%
\begin{align*}
M = & \{?m \; | \; \langle \psi^p_{n+1}, \texttt{rwr:hasAttributes}, ?x \rangle \in \Psi \\
       & \; \wedge \; \langle ?x, \texttt{rwr:hasAttribute}, ?y \rangle \in \Psi \\
       & \; \wedge \; \langle ?y, \texttt{rdf:type}, \texttt{rwr:Is} \rangle \in \Psi \\
       & \; \wedge \; \langle ?y, \texttt{rwr:steps}, ?m \rangle \in \Psi \},
\end{align*}
%%%
then
%%%
\begin{align*}
O(p)_{n+1} = \bigcup_{m \in M} g^p_{n-m} \; ,
\end{align*}
%%%
where $O(p)_{n+1} \subseteq V$ and $g^p_{n+1} \in O(p)_{n+1}$. Again, unless $O(p)_{n+1} = \emptyset$, one of the vertices in $O(p)_{n+1}$ must be $p$'s location in $G^n$ at $n+1$. 

\begin{figure}[h!]
	\centering
	\includegraphics[width=0.3\textwidth]{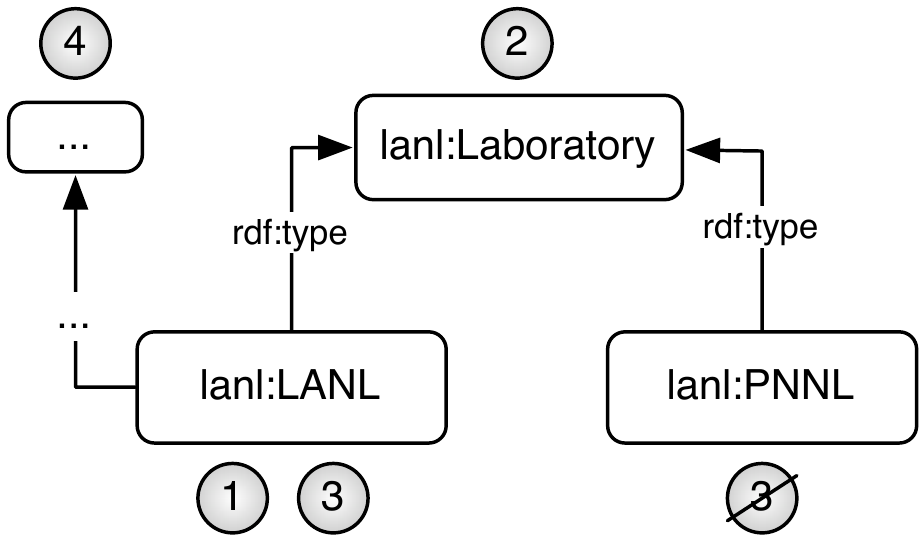}
	 \caption{An example situation for the \texttt{rwr:Is} attribute \label{fig:is-example}}
\end{figure}

The \texttt{rwr:Is} attribute is useful when $p$ must search particular properties of a vertex and later return to the original vertex. For instance, imagine the triple $\langle \texttt{lanl:LANL}, \texttt{rdf:type}, \texttt{lanl:Laboratory} \rangle \in G^n$ as depicted in Figure \ref{fig:is-example}, where the numbered circles represent the $p$'s location at particular time steps $n$. Assume that $p$ is at the \texttt{lanl:LANL} vertex at $n=1$ and $p$ must check to determine if \ttt{lanl:LANL} is, in fact, a \ttt{lanl:Laboratory}. In order to do so, $p$ must traverse the \texttt{rdf:type} predicate to arrive at \texttt{lanl:Laboratory} at $n=2$. At $n=3$, $p$ should return to the original \texttt{lanl:LANL} vertex. Without the \texttt{rwr:Is} attribute, $p$ has the potential for choosing some other \texttt{lanl:Laboratory}, such as \texttt{lanl:PNNL}.  Once back at \texttt{lanl:LANL}, it is apparent that \texttt{lanl:LANL} is a \texttt{lanl:Laboratory} and $p$ can move to some other vertex at $n=4$.
%%%

\subsection{The \ttt{rwr:Traversal} Rule}

The \texttt{rwr:Travere} rule allows the random walker $p$ to traverse to a new \texttt{rwr:Context} in $\Psi$ and a new vertex in $V$. If there exists some \texttt{rwr:Context} $\phi$ with the \texttt{rwr:Traverse} rule $t$, then when $g^p_n = a$ and $\psi^p_n = \phi$, the probability of $p$ traversing some outgoing triple from $a$ or some incoming triple to $a$ is $\frac{1}{|\Gamma(a,p)|}$, where if
%%%
\begin{align*}
	Y_\text{out} = & \{?y \; | \; \langle t, \texttt{rdfs:hasEdge}, ?y \rangle \in \Psi \\
	       & \; \wedge \; \langle ?y, \texttt{rdf:type}, \texttt{rwr:OutEdge} \rangle \in \Psi \},
\end{align*}
%%%
\begin{align*}
	Y_\text{in} = & \{?y \; | \; \langle t, \texttt{rdfs:hasEdge}, ?y \rangle \in \Psi \\
	       & \; \wedge \; \langle ?y, \texttt{rdf:type}, \texttt{rwr:InEdge} \rangle \in \Psi \},
\end{align*}
%%%
\begin{align*}
	\Gamma^+(a,p) = \bigcup_{y \in Y_\text{out}}&\{ \langle a, ?\omega, ?b \rangle \; | \; \langle a, ?\omega, ?b \rangle \in G^n \\
				        & \; \wedge \; \langle y, \texttt{rwr:hasPredicate}, ?w \rangle \in \Psi \\
				        & \; \wedge \; ( \langle ?\omega, \texttt{rdfs:subPropertyOf}, ?w \rangle \in G^n \\ 
				        & \; \; \; \; \; \; \vee \; ?\omega = ?w) \\
				        & \; \wedge \; \langle y, \texttt{rwr:hasObject}, ?x \rangle \in \Psi \\
				        & \; \wedge \; \langle ?x, \texttt{rdf:forResource}, ?z \rangle \in \Psi \\ 
				        & \; \wedge \; (\langle ?b, \texttt{rdf:type}, ?z \rangle \in G^n \; \vee \; ?b=?z) \\
				        & \; \wedge \; (O(p)_{n+1} = \emptyset \; \vee \; ?b \in O(p)_{n+1}) \\
				        & \; \; \wedge \; ?b \not\in X(p)_{n+1}  \},
\end{align*}
%%%
\begin{align*}
	\Gamma^-(a,p) = \bigcup_{y \in Y_\text{in}}&\{ \langle ?b, ?\omega, a \rangle \; | \; \langle ?b, ?\omega, a \rangle \in G^n \\
				        & \; \wedge \; \langle y, \texttt{rwr:hasPredicate}, ?w \rangle \in \Psi \\
				        & \; \wedge \; ( \langle ?\omega, \texttt{rdfs:subPropertyOf}, ?w \rangle \in G^n \\
				        & \; \; \; \; \; \; \vee \; ?\omega = ?w) \\
				        & \; \wedge \; \langle y, \texttt{rwr:hasSubject}, ?x \rangle \in \Psi \\
				        & \; \wedge \; \langle ?x, \texttt{rdf:forResource}, ?z \rangle \in \Psi \\ 
				        & \; \wedge \; (\langle ?b, \texttt{rdf:type}, ?z \rangle \in G^n \; \vee \; ?b=?z) \\
				        & \; \wedge \; (O(p)_{n+1} = \emptyset \; \vee \; ?b \in O(p)_{n+1}) \\
				        & \; \; \wedge \; ?b \not\in X(p)_{n+1}  \},
\end{align*}
%%%
then 
%%%
\begin{equation*}
\Gamma(a,p) = \Gamma^+(a,p) \cup \Gamma^-(a,p).
\end{equation*}

At the completion of the traversal, $g^p_{n+1} = b$, $g^p_{n+1'} = \omega$, $\psi^p_{n+1} = x$, and $\psi^p_{n+1'} = w$. If the edge was chosen from $\Gamma^+(a,p)$ then $g^p_{n+1''} = +$ and $\psi^p_{n+1''} = +$. If the edge was chosen from $\Gamma^-(a,p)$ then $g^p_{n+1''} = -$ and $\psi^p_{n+1''} = -$. It is always the case that $\forall n : \psi^p_{n''} = g^p_{n''}$.

Note the relationship between $G^n$ and $\Psi$ in the definition of both $\Gamma^-(a,p)$ and $\Gamma^+(a,p)$. It is necessary that the \ttt{rwr:hasPredicate} $?w$ and the \ttt{rwr:forResource} $?z$ as defined in $\Psi$ also exist in $G^n$. It is through the \ttt{rwr:Traverse} rule that the relationship between $\Psi$ and $G^n$ is made explicit and demonstrates how  $\Psi$ constrains the path that $p$ can traverse in $G^n$.

\begin{figure}[h!]
	\centering
	\includegraphics[width=0.25\textwidth]{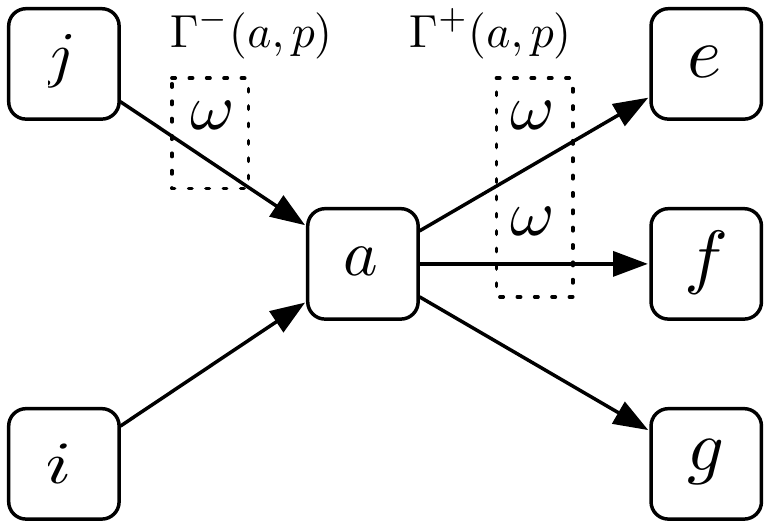}
	 \caption{ \label{fig:traverse-example}An example of the set of edges allowed for traversal by $p$ when $g^p_n = a$.}
\end{figure}

Figure \ref{fig:traverse-example} depicts an example of a traversal. In Figure \ref{fig:traverse-example}, $\Gamma_\psi^-(a,p) = \{\langle j, \omega, a \rangle \}$ and  $\Gamma_\psi^+(a,p) = \{ \langle a, \omega, e \rangle, \langle a, \omega, f \rangle \}$, where $\Gamma_\psi(i) = \{ \langle j, \omega, a \rangle, \langle a, \omega, e \rangle, \langle a, \omega, f \rangle \}$, and any one triple is selected with $\frac{1}{3}$ probability.
%%%

\subsection{The \ttt{rwr:IncrCount} and \ttt{rwr:SubmitCounts} Rules}

The purpose of the \ttt{rwr:IncrCount} and \ttt{rwr:SubmitCounts} rules is to increment the local vertex rank vector $\vpi^p$ and global vertex rank vector $\vpi$, respectively. While $\vpi^p$ is a local variable of $p$, only $\vpi$ is returned at the completion of the grammar-based random walker algorithm. The reason for $\vpi^p$ is to ensure that prior to incrementing $\vpi$, the vertices indexed by $\vpi^p$ are in a grammatically correct region of $G^n$ as determined by the grammar $\Psi$. For example, if $p$ is to index a particular \ttt{lanl:Human}, it will do so in $\vpi^p$. However, before that \ttt{lanl:Human} is considered legal according to $\Psi$, $p$ may have to check to see if the \ttt{lanl:Human} is \ttt{lanl:locatedAt} the same \ttt{lanl:University} of some previously encountered \ttt{lanl:Human}. Thus, when $p$ has submitted its $\vpi^p$ to $\vpi$, it will have guaranteed that all the appropriate aspects of its incremented vertices in $\vpi^p$ have been validated by $\Psi$. This concept will be made more salient in the example to follow in the next section.

Formally, if $\langle \phi, \texttt{rdf:type}, \texttt{rwr:Context} \rangle \in \Psi$, $\psi^p_n = \phi$, $g^p_n = i$, and $\phi$ has the \texttt{rwr:IncrCount} rule, then
%%%
\begin{equation*}
	{\vpi^p_i}_{(n+1)}  = {\vpi^p_i}_{(n)} + 1.
\end{equation*}

Next, if $g^p = i$, $\psi^p = \phi$, $\langle \phi, \texttt{rdf:type}, \texttt{rwr:Context} \rangle \in \Psi$, and $\phi$ has the \texttt{rwr:SubmitCounts} rule, then
%%%
\begin{equation*}
	{\vpi_i}_{(n+1)}= {\vpi_i}_{(n)} + {\vpi^p_i}_{(n)} \; : \; \forall i \in \vpi^p
\end{equation*}
%%%
and
%%%
\begin{equation*}
	{\vpi^p_i}_{(n+1)} = 0 \; : \; \forall i \in \vpi^p.
\end{equation*}
%%%
As stated above, once $\vpi^p$ has been submitted to $\vpi$, the values of $\vpi^p$ are set to $0$.

\subsection{The \ttt{rwr:Reresolve} Rule}

The \texttt{rwr:Reresolve} rule is a way to ``teleport" the random walker to some random vertex in $V$ and is perhaps the most complicated rule of the grammar-based random walker ontology. If there exists the \texttt{rwr:Context} $\phi$, $\psi^p_n = \phi$, $\phi$ has the \texttt{rwr:Reresolve} rule $u$, $\langle u, \texttt{rwr:probability}, ?d \rangle \in \Psi$, and $\langle u, \texttt{rwr:steps}, ?m \rangle \in \Psi$, then $p$ will have a $(d \cdot 100)$\% chance of re-resolving its path from $m$ steps ago to the current step $n$, where $d = 0.15$ in most PageRank implementations. If the random walker re-resolves, then the path from $g^p_{n-m}$ to $g^p_{n}$ is recalculated. In other words, a new path in $G^n$ is determined with respects to the \ttt{rwr:Context}s $\psi^p_{n-m}$ to $\psi^p_{n}$ such that no rules are executed and only those attributes specified by the $\texttt{rwr:obeys}$ property are respected.

For example, suppose $\psi^p_{(n-m) \rightarrow n} = (\phi_{(n-m)}, \omega_{(n-m)+1'}, \pm_{(n-m)+1''}, \ldots, \omega_{n'},  \pm_{n''}, \phi_n)$ and context $\psi^p_n$ has a \texttt{rwr:Reresolve} rule, where $\psi^p_n =  \phi_n$. If the \texttt{rwr:Reresolve} rule $\texttt{rwr:obeys}$ both the \texttt{rwr:Is} and \texttt{rwr:Not} attributes, then the grammar-based random walker $p$ will re-resolve its history in $G^n$. Thus it will recalculate $g^p_{n-m}$ to $g^p_{n}$. The set of legal re-resolved paths from $n-m$ steps ago to $n$  is denoted $Q_{(n-m),n}$. Given that the probability $d$ is met,

\begin{align*}
Q_{(n-m),n} = & \{ ( ?i, ?\omega_{(n-m)+1'},\pm_{(n-m)+1''}, ?a, \ldots, \\ 
		& \;\;\;\;\;\;\;\;\;\; ?b, ?\omega_{n'}, \pm_{n''}, ?j ) \; | \\
		& \;\;\;\;\;\; \langle \psi^p_{(n-m)}, \texttt{rwr:forResource}, ?x \rangle \in \Psi \\
		& \; \wedge \; (\langle ?i, \texttt{rdf:type}, ?x \rangle \in G^n \; \vee \; ?i=?x) \\
		& \; \wedge \; (O(p)_{(n-m)} = \emptyset \; \vee \; ?i \in O(p)_{(n-m)}) \\
		& \; \; \wedge \; ?i \not\in X(p)_{(n-m)} \\
		& \; \wedge \; (\langle ?\omega_{n'}, \texttt{rdfs:subPropertyOf}, \psi^p_{n'}  \rangle \in G^n \; \\
		& \; \; \; \; \; \; \; \vee \; \; ?\omega_{(n-m)+1'}=\psi^p_{(n-m)+1'}) \\
		& \; \wedge \; ((\pm_{(n-m)+1''} = + \\
		& \;\;\;\;\;\;\;\;\; \wedge \; (?i,?\omega_{(n-m)+1'},?a) \in G^n) \\
		& \; \; \; \; \; \;  \vee \; (\pm_{(n-m)+1''} = - \\
		& \;\;\;\;\;\;\;\;\; \wedge \; (?a,?\omega_{(n-m)+1'},?i) \in G^n)) \\
		& \; \wedge \;  \ldots \\
		& \; \wedge \; ((\pm_{n''} = + \; \wedge \; (?b,?\omega_{n'},?j) \in G^n) \\
		& \; \; \; \; \; \;  \vee \; (\pm_{n''} = - \; \wedge \; (?j,?\omega_{n'},?b) \in G^n)) \\
		& \; \wedge \; (\langle ?\omega_{n'}, \texttt{rdfs:subPropertyOf}, \psi^p_{n'}  \rangle \in G^n \; \\
		& \; \; \; \; \; \; \; \vee \; ?\omega_{n'} = \psi^p_{n'} ) \\
		& \; \wedge \; \langle \psi^p_{n}, \texttt{rwr:forResource}, ?y \rangle \in \Psi \\
		& \; \wedge \; (\langle ?j, \texttt{rdf:type}, ?y \rangle \in G^n \; \vee \; ?j=?y) \\
		& \; \wedge \; (O(p)_{n} = \emptyset \; \vee \; ?j \in O(p)_{n}) \\
		& \; \; \wedge \; ?j \not\in X(p)_{n}  \}.
\end{align*}
%%%
The probability of $p$ choosing some re-resolved path $q \in Q_{(n-m),n}$ is $\frac{1}{Q_{(n-m),n}}$, where $g^p_{k''} = q_{k''}$, $g^p_{k'} = q_{k'}$, and $g^p_{k} = q_{k}$ for all $k$ such that $m \leq k \leq n$.

While the above equation is perhaps notationally tricky, it has a relatively simple meaning. In short, $p$ must recalculate (or re-resolve) its path from $m$ step ago to the present step $n$. This recalculation must follow the exact same grammar path denoted in $\psi^p$. Thus, if from $m$ to $n$, $p$ had ensured that its current vertex is a \ttt{lanl:Human} that is \ttt{lanl:locatedAt} \ttt{lanl:Laboratory} then when $p$ ``teleports", the new vertex at $n$ will be guaranteed to also be a \ttt{lanl:Human} that is \ttt{lanl:locatedAt} a \ttt{lanl:Laboratory}.

If there are no rank sinks, this rule guarantees a strongly connected network; any vertex can be reached by any other vertex in the grammatically correct region of $G^n$. However, note that rank sinks are remedied by the next rule.

\subsection{The Empty Rule}

Random walker halting occurs when $p$ arrives at some \ttt{rwr:Context} where no rule exists or there are no more rules to execute (e.g.~when a \ttt{rwr:Traverse} rule does not provide any transition edges -- $\Gamma(a,p) = \emptyset$). At halt points, a new random walker with an empty $\vpi^p$ and no $G^n$ or $\Psi$ history (i.e.~$|g^p|=0$ and $|\psi^p|=0$), enters $G^n$ at some \ttt{rwr:EntryContext} $\phi$ in $\Psi$ and some $i \in s(V \; | \; \phi)$. The new random walker executes the grammar. Note that the global rank vector $\vpi$ remains unchanged.

The combination of the empty rule and the \texttt{rwr:Reresolve} rule are necessary to ensure that $\vpi$ is a stationary distribution. Both rules are used in conjunction to support grammar-based PageRank calculations.

In order to demonstrate the aforementioned ideas, the next section presents a particular grammar instance developed for a scholarly network ontology and instance.

\section{A Scholarly Network Example}

\begin{figure}[h!]
	\centering
	\includegraphics[width=0.495\textwidth]{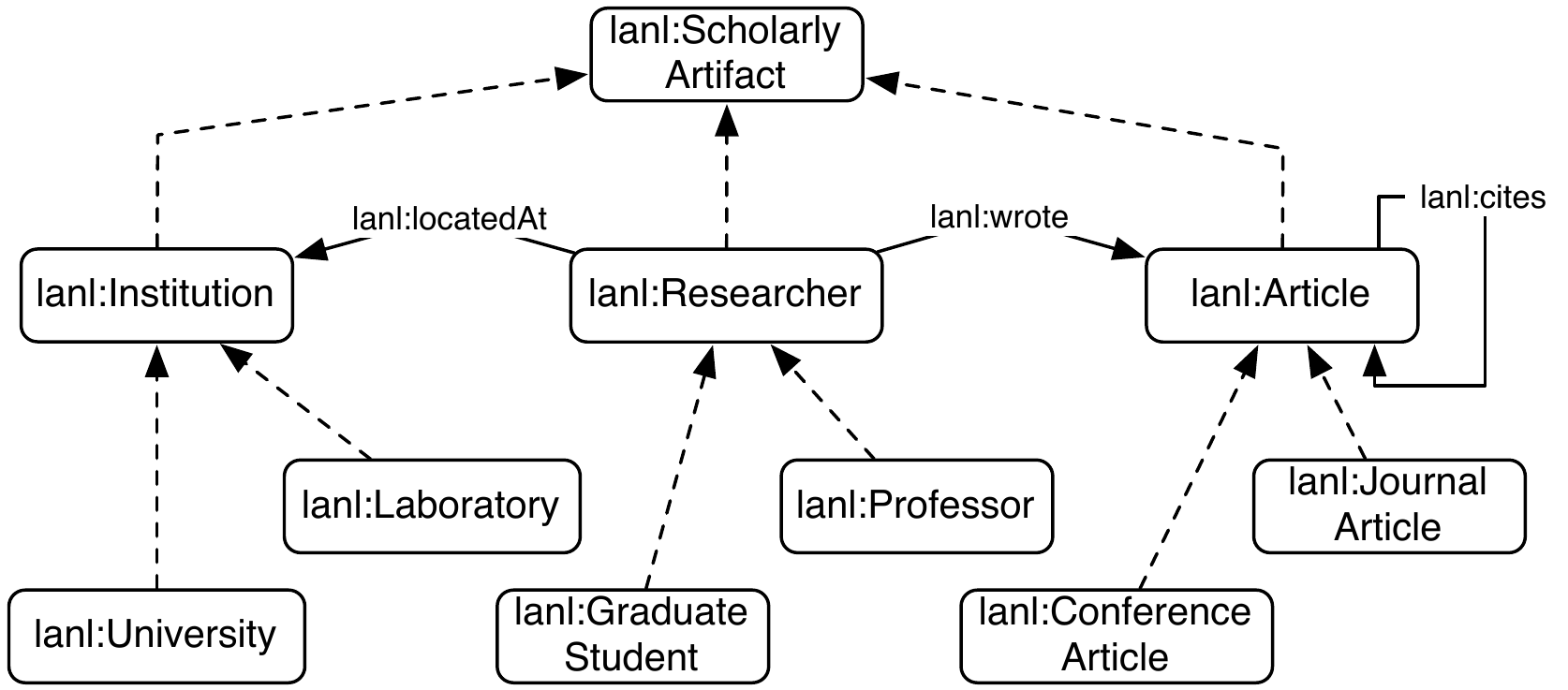}
	 \caption{An example scholarly ontology \label{fig:schol-1}}
\end{figure}

This section will demonstrate the application of grammar-based random walkers to a scholarly semantic network denoted $G^n$. Figure \ref{fig:schol-1} diagrams the ontology of $G^n$ where the tail of the edge is the \ttt{rdfs:domain} and the head of the edge is the \ttt{rdfs:range}. The dashed lines represent the \texttt{rdfs:subClassOf} relationship. This ontology represents the relationships between \ttt{lanl:Institution}s, \ttt{lanl:Researcher}s, \ttt{lanl:Article}s, and their respective children classes.
%%%

The first example calculates the stationary distribution of the subset of $G^n$ that is semantically equivalent to the coauthorship network resulting from \ttt{lanl:ConferenceArticle}s written by \ttt{lanl:Researcher}s that are \ttt{lanl:locatedAt} a \ttt{lanl:University} only. The second example presents a grammar for calculating the stationary distribution over all vertices in a semantic network irrespective of the edge labels (i.e.~an unconstrained grammar). The second example is equivalent to running the single-relational implementation of PageRank on a semantic network.

\subsection{Conference Article Co-Authorship Grammar $\Psi_\text{coaut}$}

\begin{figure}[h!]
	\centering
	\includegraphics[width=0.48\textwidth]{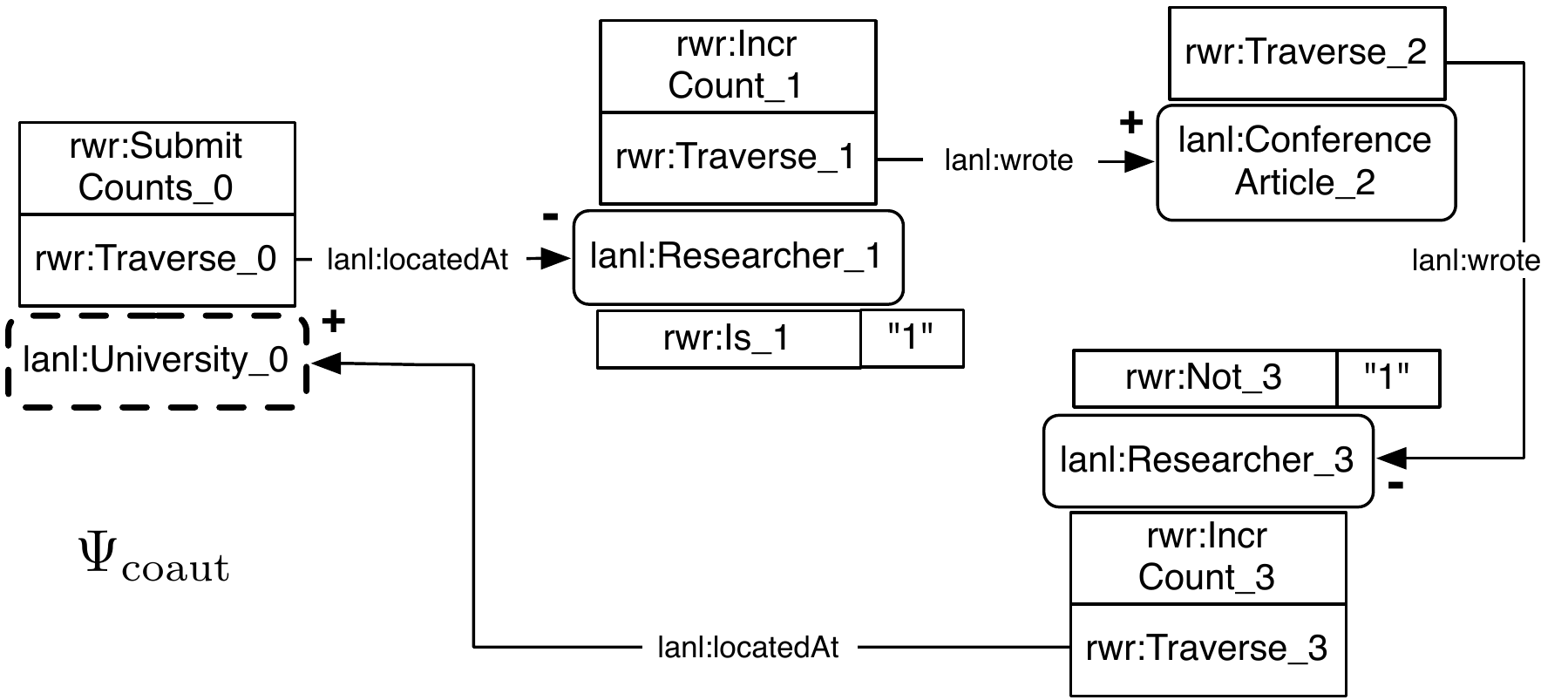}
	 \caption{\label{fig:coaut}A grammar to calculate eigenvector centrality on a conference article coauthorship network of university researchers.}
\end{figure}

Let $\Psi_{\text{coaut}}$ denote the grammar for generating a $\vpi$ for the subset of $G^n$ that is semantically equivalent to the coauthorship network resulting from \ttt{lanl:ConferenceArticle}s for all \ttt{lanl:Researcher}s from a \ttt{lanl:University}. $\Psi_{\text{coaut}}$ is diagrammed in Figure \ref{fig:coaut} where, for the sake of convenience, the context names, without the \_\#, denote the \texttt{rdfs:Resource} pointed to by the \texttt{rwr:forResource} property of the respective \texttt{rwr:Context}. The bolded $+$ or $-$ on the edges denotes whether the \ttt{rwr:Edge} is an \texttt{rwr:OutEdge} or \ttt{rwr:InEdge}, respectively. The dashed square represents an \texttt{rwr:EntryContext}. The stack of rules for each \texttt{rwr:Context} denotes the \texttt{rdf:Seq} of rules ordered from top to bottom and \texttt{rwr:Context} attributes are also stacked (in no particular order) with their respective \texttt{rwr:Context}.
%%%

A single grammar-based random walker $p \in P$ will begin its journey in $G^n$ at some vertex $i \in s(V \; | \; \texttt{lanl:University\_0})$, where
%%%
\begin{align*}
s(V \; | \; & \texttt{lanl:University\_0}) = \\
&  \{ ?i \; | \; \langle ?i, \texttt{rdf:type}, \texttt{lanl:University} \rangle \in G^n \}
 \end{align*}
 %%%
and the $i \in s(V \; | \; \texttt{lanl:University\_0})$ is chosen with probability $\frac{1}{|s(V \; | \; \texttt{lanl:University\_0})|}$. After a vertex in $s(V \; | \; \texttt{lanl:University\_0})$ is chosen, $g^p_0 = i$ and $\psi^p_0 = \texttt{lanl:University\_0}$. There are $2$ sequentially ordered rules at  \texttt{University\_0}: \texttt{rwr:SubmitCounts\_0} and \texttt{rwr:Traverse\_0}. The first rule has no effect on $\vpi$ or $\vpi^p$ because for all $i$ ${\vpi^p_{i}}_{(0)} = 0$. The \texttt{rwr:SubmitCounts\_0} rule is important on the next time around $\Psi_{\text{coaut}}$. With the \texttt{rwr:Traverse\_0} rule, $p$ randomly chooses a single vertex $w$ in
%%%
\begin{align*}
W =& \{ ?w \; | \; \langle ?w, \texttt{lanl:locatedAt}, i \rangle \in G^n \\
       & \; \wedge \;  \langle ?w, \texttt{rdf:type}, \texttt{lanl:Researcher} \rangle \in G^n \},
\end{align*}
%%%
where $\texttt{rwr:Is\_1}$ requires that $g^p_{1} = g^p_{-1}$ and $g^p_{-1} = \emptyset$ (i.e.~$O(p)_1 = \emptyset$). The \texttt{rwr:Is\_1} attribute is important the second time around $\Psi_{\text{coaut}}$. 

At time step $1$, $g^p_{1} = w$ and $\psi^p_{1} = \texttt{lanl:Researcher\_1}$. \texttt{Researcher\_1} has the \texttt{rwr:IncrCount\_1} rule and thus, ${\vpi^p_{w}}_{(1)} = 1$. After the \ttt{rwr:IncrCount\_1} rule is executed, $p$ will execute the \ttt{rwr:Traverse\_1} rule. The random walker $p$ will randomly choose some $x$ in
%%%
\begin{align*}
X= &  \{ ?x \; | \; \langle w, \texttt{lanl:wrote}, ?x \rangle \in G^n \\
      & \; \wedge \; \langle  ?x, \texttt{rdf:type}, \texttt{lanl:ConferenceArticle} \rangle \in G^n \}.
\end{align*}
%%%
If $x$ is properly resolved, then $g^p_{2} = x$ and $\psi^p_{2} = \texttt{lanl:ConferenceArticle\_2}$. However, if $w$ has not written a \ttt{lanl:ConferenceArticle}, then $x = \emptyset$. At which point, the \texttt{rwr:Traverse\_1} rule fails and $(i, \texttt{lanl:locatedAt}, -,  w)$ is an ungrammatical path in $G^n$ according to $\Psi_{\text{coaut}}$. If $x = \emptyset$, a new random walker (i.e.~a $p$ with no history and zero $\vpi^p$) randomly chooses some entry point into $\Psi_{\text{coaut}}$ and $G^n$ and the process begins again. If, on the other hand, $w$ has written some \ttt{lanl:ConferenceArticle} $x$, then $p$ will randomly select a $y$ in
%%%
\begin{align*}
Y =& \{ ?y \; | \; \langle ?y, \texttt{lanl:wrote}, x \rangle \in G^n \\
      & \; \wedge \; \langle ?y, \texttt{rdf:type}, \texttt{lanl:Researcher} \rangle \in G^n \\
      & \; \; \wedge \; ?y \; \neq \; w \}.
\end{align*}

Note the role of the \texttt{rwr:Not\_3} property in \texttt{Researcher\_3}. \texttt{rwr:Not\_3} guarantees that the $x$ \ttt{lanl:ConfereneArticle} was written by two or more \ttt{lanl:Researcher}s and that only those \ttt{lanl:Researcher}s that are not $w$ are selected since $X(p)_3 = \{w\}$. Semantically, this ensures that the subset of $G^n$ that is traversed is a coauthorship network. If $y = \emptyset$, then $(i, \texttt{lanl:locatedAt}, -, w, \texttt{lanl:wrote}, +, x)$ is an ungrammatical path with respects to $\Psi_{\text{coaut}}$. If  $y \neq \emptyset$, then $g^p_{3} = y$, $\psi^p_3 = \texttt{Researcher\_3}$, and ${\vpi^p_{y}}_{(3)} = 1$. Finally, because of the \ttt{rwr:Traverse\_3} rule, $p$ randomly selects some $z$ in
%%%
\begin{align*}
Z  =& \{?z \; | \;  \langle y, \texttt{lanl:locatedAt}, ?z \rangle \in G^n \\
       & \; \wedge \; \langle ?z, \texttt{rdf:type}, \texttt{lanl:University} \rangle \in G^n \}.
\end{align*}

Thus, $g^p_4 = z$ and $\psi^p_4 = \texttt{University\_0}$. At this point in time, $g^p = (i,$ \texttt{lanl:locatedAt}, $-, w$, \texttt{lanl:wrote}, $+, x$, \texttt{lanl:wrote}, $-, y$, \texttt{lanl:locatedAt}, $+, z)$ and $g^p$ is a $\Psi_{\text{coaut}}$-correct and $w$ and $y$ are indexed by $\vpi$. The \texttt{rwr:SubmitCounts\_0} rule ensures that ${\vpi_{w}}_{(4)} = {\vpi^p_{w}}_{(4)}$ and ${\vpi_{y}}_{(4)} = {\vpi^p_{y}}_{(4)}$. Finally, when \ttt{rwr:SubmitCounts\_0} has completed, ${\vpi^p_{w}}_{(4)} = {\vpi^p_{y}}_{(4)} = 0$. This process continues until the ratio between the counts in $\vpi$ converge.

At $n=5$, the \ttt{rwr:Is\_1} rule is important to ensure that, after checking if the $y$ \ttt{rwr:Researcher} is \ttt{rwr:locatedAt} a \ttt{rwr:University}, $p$ return to $y$ before locating a \ttt{rwr:ConferenceArticle} written by $y$ and continuing its traversal through the implicit coauthorship network in $G^n$ as defined by $\Psi_{\text{coaut}}$.

What is provided by $\vpi$ is the number of times a particular vertex in $V$ has been visited over a given number of time steps $n$. If vertex $i \in V$ was visited $\vpi_i$ times then the probability of observing a random walker at $i$ is $\frac{n}{\vpi_i}$. However, given that $\sum_{i \in V} \vpi_i \leq n$ because other vertices not indexed by $\vpi$ exist on a $\Psi_{\text{coaut}}$-correct path of $G^n$, the probability of the random walker being at vertex $i$ when observing only those vertices indexed by $\vpi$ is
%%%
\begin{equation*}
	\vpi_i' = \frac{\vpi_i}{\sum_{j \in V} \vpi_j} \; : \; i \in V.
\end{equation*} 
%%%
Thus,
%%%
\begin{equation*}
	\sum_{i \in V} \vpi_i' = 1.
\end{equation*}

This step is called the normalization of $\vpi$ and is necessary for transforming the number of times a vertex in $V$ is visited into the probability that the vertex is being visited at any one time step. When ${||\vpi'_{(n)} - \vpi'_{(m)}||}_2 \leq \epsilon$, where $m < n$ and $m$ and $n$ are consecutive $\vpi$ update steps (i.e.~consecutive \texttt{rwr:SubmitCounts}), $\vpi$ has converged to a range acceptable by the $\epsilon \in \mathbb{R}$ provided argument.

However, $\vpi$ may never converge if the $p$-traversed subset of $G^n$ is not strongly connected. For instance, let the triple list $A^n$ be defined as
%%%
\begin{align*}
A^n =& \{ \langle ?i, \texttt{lanl:coauthor}, ?y \rangle \; | \\
	& \; \wedge \; \langle ?w, \texttt{rdf:type}, \texttt{lanl:University} \rangle \in G^n \\
          & \; \wedge \; \langle ?i, \texttt{lanl:locatedAt}, ?w \rangle \in G^n \\
          & \; \wedge \; \langle ?i, \texttt{lanl:wrote}, ?x \rangle \in G^n \\
          & \; \wedge \; \langle ?x, \texttt{rdf:type}, \texttt{lanl:ConferenceArticle} \rangle \in G^n \\
          & \; \wedge \; \langle ?y, \texttt{lanl:wrote}, ?x \rangle \in G^n \\
          & \; \wedge \; \langle ?y, \texttt{lanl:locatedAt}, ?z \rangle \in G^n \\
          & \; \wedge \; \langle ?z, \texttt{rdf:type}, \texttt{lanl:University} \rangle \in G^n \\
          & \; \wedge \; ?i \neq ?y \}.
\end{align*} 
%%%
Furthermore, let $V^*$ denote the set of unique \ttt{lanl:Researcher} vertices in $A^n$ and $\mathbf{A} \in \mathbb{R}^{|V^*| \times |V^*|}$ be a weighted adjacency matrix where
%%%
\begin{equation*}
\mathbf{A}_{i,y} = 
	\begin{cases}
		\frac{1}{|\Gamma^+(i)|} & \text{if } \langle i, \texttt{lanl:coauthor}, y \rangle \in A^n \\
		\frac{1}{|V^*|} & \text{if } |\Gamma^+(i)| = 0.
	\end{cases}
\end{equation*}

\begin{figure}[h!]
	\centering
	\includegraphics[width=0.2\textwidth]{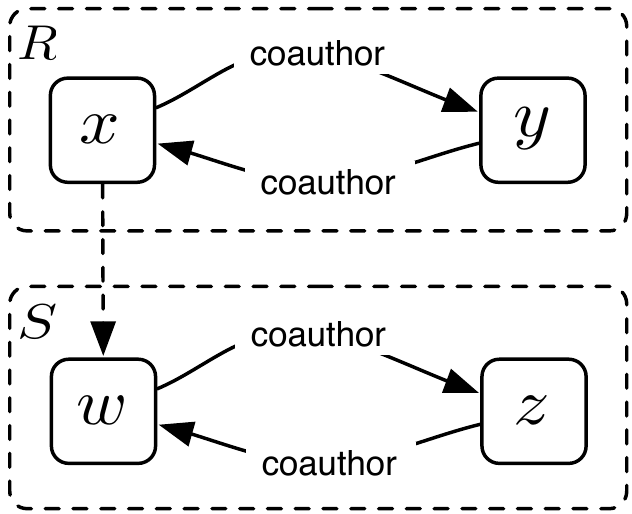}
	 \caption{ \label{fig:coaut-example}Teleportation required for connecting isolated components.}
\end{figure}

If $\mathbf{A}\vpi' = \lambda\vpi'$ where $\lambda$ is the largest eigenvalue of the eigenvectors of $\mathbf{A}$, then $\vpi'$ is the stationary distribution of $\mathbf{A}$ and thus, the $p$-traversed subset of $G^n$ given $\Psi_{\text{coaut}}$ is strongly connected. However, most coauthorship networks are not strongly connected \cite{coauth:liu2005} and therefore, $\vpi'$ may not be a stationary distribution. For example, there may exists some \ttt{lanl:University} denoted $R$ and \ttt{lanl:locatedAt} $R$ are only two \ttt{lanl:Researcher}s, $x$ and $y$, that have a coauthor relationship with respects to a particular \ttt{lanl:ConferenceArticle}. If the random walker $p$ happens to enter $G^n$ at $x$, then the random walker will never leave the $x/y$ component. However, some new \ttt{lanl:Researcher}, and therefore some new \ttt{lanl:University}, can be introduced into the problem by re-resolving the \ttt{lanl:ConferenceArticle} uniting $x$ and $y$ such that $p$ teleports to some new researcher $w$ at some other \ttt{lanl:University} $S$. This example is depicted in Figure \ref{fig:coaut-example}, where the dashed line represents a teleportation by $p$. This teleportation introduces the artificial relationship that $x$ coauthored with $w$. Thus, when there exists a non-zero probability of teleportation at every vertex in $V^*$, the coauthorship network becomes strongly connected.
%%%

\begin{figure}[h!]
	\centering
	\includegraphics[width=0.48\textwidth]{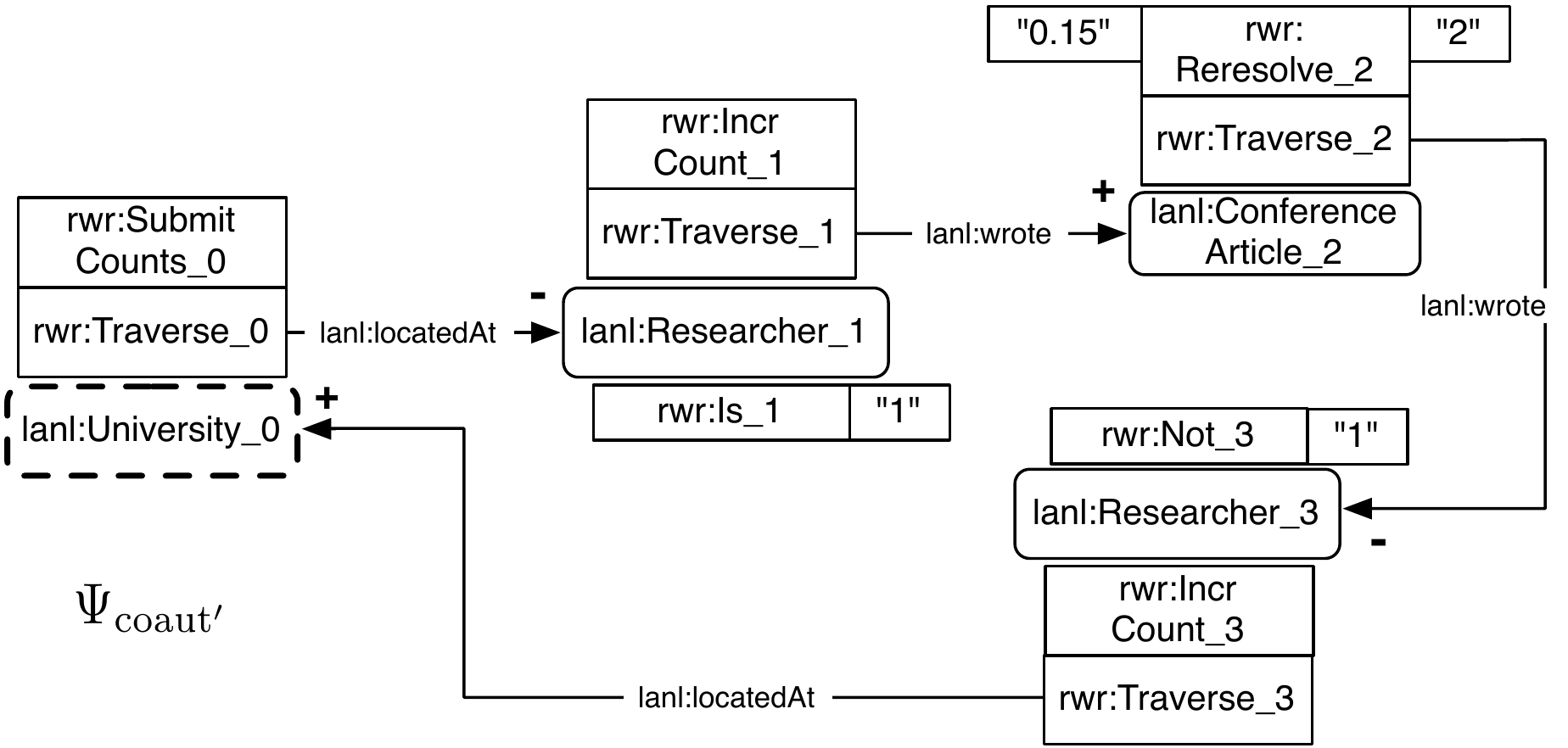}
	 \caption{ \label{fig:coaut2}A grammar to calculate PageRank on a conference article coauthorship network of university researchers.}
\end{figure}

In order to guarantee a strongly connected network, it is possible to simulate the behavior of randomly choosing some new entry point with probability $\delta \in (0,1] \;$Êas an analogy to the method of inducing strong connectivity in \cite{page98pagerank}. The $\texttt{rwr:Reresolve}$ rule is introduced to $\Psi_{\text{coaut}}$ at \texttt{ConferenceArticle\_2} where \texttt{rwr:Reresolve\_2} has a $\delta = 0.15$, a \texttt{rwr:steps} of $m = 2$, and does not \ttt{rwr:obey} any \texttt{rwr:Context} attributes. $\Psi_{\text{coaut'}}$ is diagrammed in Figure \ref{fig:coaut2}, where the \texttt{"0.15"} literal is the object of the triple $\langle \texttt{rwr:Reresolve\_2}, \texttt{rwr:probability}, \texttt{"0.15"} \rangle \in \Psi_{\text{coaut'}}$ and the \texttt{"2"} literal is the object of triple $\langle \texttt{rwr:Reresolve\_2}, \texttt{rwr:steps}, \texttt{"2"} \rangle \in \Psi_{\text{coaut'}}$.\\
%%%

With respects to $G^n$, every time random walker $p$ encounters the \texttt{rwr:ConferenceArticle\_2} context, it has a 15\% chance of teleporting to some new \texttt{lanl:ConferenceArticle} $i$ in $V$ such that
%%%
\begin{align*}
Q_{n-2, n} =& \{ (?w,?x, -, ?y,?z, +, ?i) \; | \\
      & \; \wedge \; \langle ?w, \texttt{rdf:type}, \texttt{lanl:University} \rangle \in G^n \\ 
      & \; \; \wedge \; ?x = \texttt{lanl:locatedAt}  \\ 
      & \; \wedge \; \langle ?y, \texttt{rdf:type}, \texttt{Researcher} \rangle \in G^n \\ 
      & \; \; \wedge \; ?z = \texttt{lanl:wrote} \\ 
      & \; \wedge \; \langle ?i, \texttt{rdf:type}, \\
      & \; \; \; \; \; \; \; \; \; \; \; \; \; \; \; \texttt{lanl:ConferenceArticle} \rangle \in G^n \\
      & \; \wedge \; \langle ?y, ?x, ?w \rangle \in G^n \\
      & \; \wedge \; \langle ?y, ?z, ?i \rangle \in G^n \}.
\end{align*}
%%%
and a new path $q \in Q_{n-2,n}$ is chosen with probability $\frac{1}{|Q_{n-2,n}|}$. If $q = (w,x,-,y,z,+,i)$,
%%%
\begin{equation*}
	g^p_{(n-2) \rar n} =
	\begin{cases}
		g^p_{(n-2) \rar n} & \text{with probability } 1-d \\
		q_{0 \rar 2}	    & \text{with probability } d.
	\end{cases} 
\end{equation*}

The \texttt{rwr:Reresolve} rule guarantees that any conference publishing researcher is reachable by any other conference publishing university researcher and thus, the coauthorship network of conference publications by university researchers is strongly connected. Theoretically, the $\texttt{rwr:Reresolve\_2}$ rule ensures that there exists some hypothetical triple list $B^n$, such that
%%%
\begin{equation*}
B^n = \{ \langle ?i, \texttt{lanl:teleport}, ?j \rangle \; | \; ?i,?j \in V^* \},
\end{equation*}
%%%
where $V^*$ is the set of \ttt{lanl:Researcher}s from $A^n$. Let $\mathbf{B} \in \mathbb{R}^{|V^*| \times |V^*|}$ be a weighted adjacency matrix where for any entry in $\mathbf{B}$, $\mathbf{B}_{i,j} = \frac{1}{|V^*|}$. $\Psi_\text{coaut'}$ is equivalent to computing $\vpi'$ for $\mathbf{C}$ where $\mathbf{C} = \delta \mathbf{A} + (1-\delta) \mathbf{B}$ and $\delta = 0.85$. Therefore, $\vpi'$ generated from $\Psi_\text{coaut'}$ is a stationary distribution.

The eigenvector centrality or PageRank of the network could have been calculated by extracting the appropriate \ttt{lanl:Researcher} vertices from $V$ and generating the implicit \ttt{lanl:ConferenceArticle} coauthorship edge between them. This was done with the network $A^n$ and its ``teleporation" network $B^n$, where $\mathbf{A}$ and $\mathbf{B}$ are the respective adjacency matrices representations of these networks. In this sense, the single-relational eigenvector centrality or PageRank algorithm would generate the same results. However, the grammar-based random walker algorithm is different than the ``isolation-based" method. In the grammar-based method, there is no need to generate (i.e.~make explicit) the implicit single-relational subset of $G^n$ and thus, create another data structure; the same $G^n$ can be used for different eigenvector calculations without altering it. Thus, multiple different grammars can be running in parallel on the same data set (on the same triple-store). For more complex grammars that involve \ttt{rwr:Is} and \ttt{rwr:Not} constraints over multiple cycles of a grammar, the query to isolate the sub-network becomes increasingly long as recursions cannot be expressed in the standard RDF query language SPARQL \cite{sparql:prud2004}. 

\subsection{Simulating Single-Relational PageRank on a Semantic Network}

\begin{figure}[h!]
	\centering
	\includegraphics[width=0.35\textwidth]{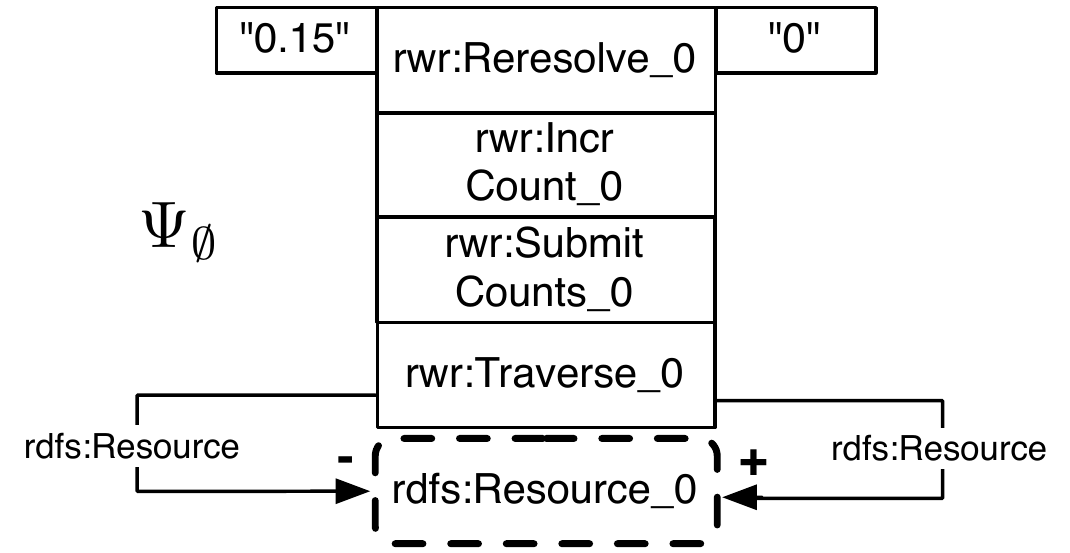}
	 \caption{ \label{fig:unlabel}A grammar to calculate an undirected single-relational network PageRank on a semantic network.}
\end{figure}

The grammar depicted in Figure \ref{fig:unlabel} is denoted $\Psi_\emptyset$ and is the grammar that calculates $\vpi$ on any semantic network without consideration for edge directionality nor edge labels. Thus, this grammar is not constrained to the ontology of the semantic network and can be applied to any $G^n$ instance. Furthermore, the \texttt{rwr:Reresolve} rule guarantees that all vertices are reachable from all other vertices. Note that this grammar ensures that all vertices in $V$ are $\Psi_\emptyset$-correct. The presented grammar is equivalent to executing PageRank on an undirected single-relational representation of a semantic network.
%%%

\section{Analysis}

What has been presented thus far is an ontology for instantiating an $G^n$ specific grammar, the formalization of the rules and attributes that must be respected by a grammar-based random walker, and an eigenvector centrality and PageRank example involving a semantic scholarly network. This section will briefly discuss the various permutations of $G^n$ that are traversed by a grammar-based random walker.

As stated previously, only a subset of the complete semantic network $G^n$ is traversed by any $p \in P$. Let $G^{\psi} \subseteq G^n$ denote the graph traversed by $p$ according to $\Psi$. It is noted that only a subset of $G^{\psi}$ is considered $\Psi$-correct (i.e.~grammatically correct according to $\Psi$). If $p$ is unable to submit its $\vpi^p$ to the global vertex vector $\vpi$, then $p$ has taken an ungrammatical semantic path in $G^n$. On the other hand, if $p$ contributes its $\vpi^p$ to $\vpi$, $p$ has taken a grammatical semantic path (i.e.~a $\Psi$-correct path). Let $G^{\psi+} \subseteq G^{\psi}$ denote the subset of $G^\psi$ that is grammatically correct according to $\Psi$.

\begin{definition}[The $\Psi$-Correct Paths of $G^{\psi+}$] \label{def:psicorrect}
The path $g^p_{m \rightarrow n}$ in $G^n$ is considered grammatically correct with respects to $\Psi$ if and only if $\psi^p_{m}$ is an \texttt{rwr:EntryContext} or an \ttt{rwr:Context} with an \ttt{rwr:SubmitCounts} rule, $\psi^p_{n}$ is an \texttt{rwr:Context} with an \texttt{rwr:SubmitCounts} rule, and there exist some \texttt{rwr:IncrCount} rule at time $k$, such that $m \leq k \leq n$. The set of all grammtically correct paths form the semantic network $G^{\psi+}$, where $G^{\psi+} \subseteq G^{\psi} \subseteq G^n$.
\end{definition}

The grammatically correct path $g^p_{m \rightarrow n}$ ensures that some vertex in $g^p_{m \rightarrow n}$ was validated by the grammar $\Psi$ and indexed by $\vpi$.
 
\begin{figure}[h!]
	\centering
	\includegraphics[width=0.225\textwidth]{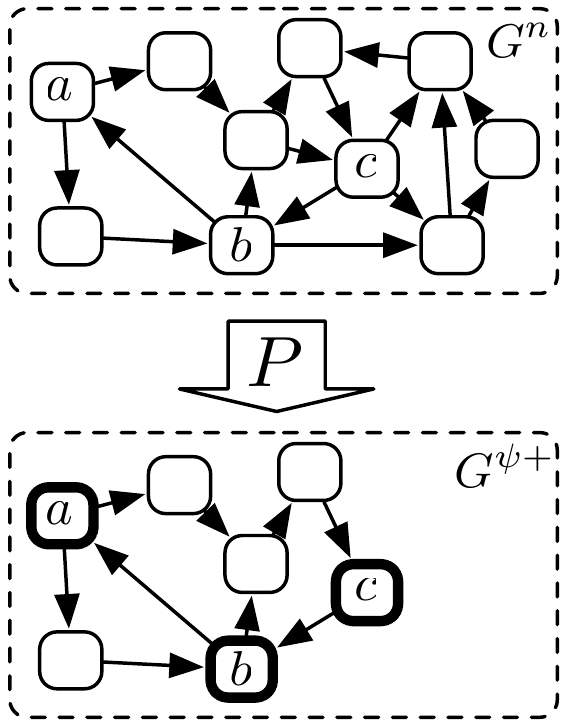}
	 \caption{\label{fig:reduce-g}$G^{\psi+}$ as the $\Psi$-correct subset of $G^n$.}
\end{figure}

Figure \ref{fig:reduce-g} demonstrates a subset of $G^n$ that is traversed by $P$ to generate $G^{\psi+}$, where the bold labeled vertices are those indexed by $\vpi$.
%%%

\begin{figure}[h!]
	\centering
	\includegraphics[width=0.225\textwidth]{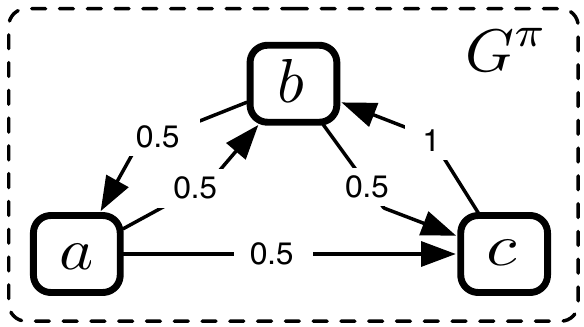}
	 \caption{\label{fig:reduce-g2}$G^{\vpi}$ as the implied network of $G^{\psi+}$.}
\end{figure}

Note that the vertices indexed by $\vpi$ are not necessarily all of the vertices encountered by the random walkers in $G^{\psi+}$. Similar to the coauthor example presented previous, while a $p \in P$ traverses vertices of type \ttt{lanl:Article}, \ttt{lanl:University}, and \ttt{lanl:Researcher}, only \ttt{lanl:Resercher} vertices are index by $\vpi$. Thus, those vertices indexed by $\vpi$ form an ``implied" network. The $G^{\psi+}$ represented in Figure \ref{fig:reduce-g} has the implied network $G^{\vpi}$ as diagrammed in Figure \ref{fig:reduce-g2}. The probabilities on the edges are given by branches between the respective vertices in $G^{\psi+}$.
%%%

\begin{theorem}
If $G^{\psi+}$ is strongly connected and aperiodic, then $\vpi'$ is a stationary probability distribution.
\end{theorem}
%%%
\emph{Proof.} If $G^{\psi+}$ is strongly connected, then every vertex in $G^{\psi+}$ is reachable from any other vertex. Given that $\vpi$ indexes a subset of the vertices in $G^{\psi+}$ and the vertices in $G^{\vpi}$ are reachable by means of the edges in $G^{\psi+}$, then the vertices indexed by $\vpi$ are strongly connected. Thus, the normalization of $\vpi$, $\vpi'$, is a stationary probability distribution. \qed

\begin{figure}[h!]
	\centering
	\includegraphics[width=0.25\textwidth]{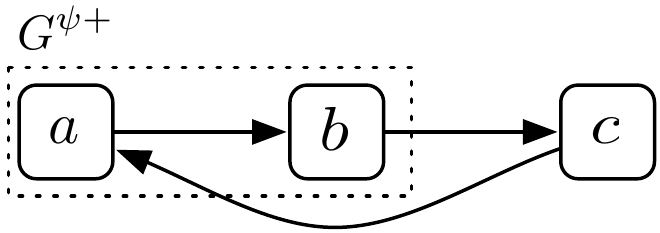}
	 \caption{\label{fig:lemma}A strongly connected $G^n$ does not guarantee a strongly connected $G^{\psi+}$.}
\end{figure}

Note that the above does not generalize to $G^n$. If $G^n$ is strongly connected, that does not guarantee that the grammar  will permit the grammar-based random walker $p$ to traverse a subset of $G^n$ that is strongly connected. For example, imagine the network $G^n$ depicted in Figure \ref{fig:lemma}. Even if $G^n$ is a strongly connected network, the $\Psi$-correct subgraph of $G^n$ traversed may not be.
%%%

\begin{figure}[h!]
	\centering
	\includegraphics[width=0.3\textwidth]{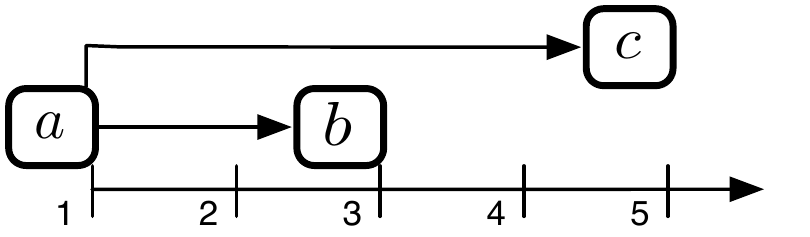}
	 \caption{\label{fig:grammar-process}The variability of index delay times for $G^\vpi$.}
\end{figure}

Finally, if the path distance between the vertices in $G^{\vpi}$ is equal in $G^{\psi+}$, then $\vpi'$ is the primary eigenvector of the $G^{\vpi}$. However, this is not always the case. Figure \ref{fig:grammar-process} demonstrates that the timing between indexing the different vertices in the network diagrammed in Figure \ref{fig:reduce-g} is different for different paths chosen by $p$.
%%%

\begin{theorem}
If the paths in $G^{\psi+}$ between the vertices indexed by $\vpi$ are of equal length, then $\vpi'$ is the primary eigenvector of $G^\vpi$.
\end{theorem}
%%%
\emph{Proof.}
If the path lengths in $G^{\psi+}$ between the vertices index by $\vpi$ are of equal length, then the intervening non-$\vpi$ vertices in $G^{\psi+}$ can be removed without interfering with the relative timing of respective increments to the vertices in $\vpi$. Given this network manipulation, a single-relational eigenvector centrality algorithm on the single-relational network $G^\vpi$ would yield $\vpi'$. Thus, $\vpi'$ is the primary eigenvector of the network $G^\vpi$. \qed 

\section{Conclusion}

There is much disagreement to the high-level meaning of the primary eigenvector of a network. $\vpi$ has been associated with concepts such as ``prestige"', ``value", ``importance", etc. For Markov chain analysis, when vertices represent states of a system,  the meaning is clear; $\vpi$ defines the probability that at some random time $n$, the system $G^1$ will be at some particular state in $V$, where more ``central" states (i.e.~those with a higher $\vpi$ probability) are more likely to been seen.

However, the application of $\vpi$ to more abstract concepts of centrality such as ``value" has been applied in the area of the web citation network. If the web is represented as a Markov chain, then $\vpi_i$ defines the probability that some random web surfer will be at a particular web page $i$ at some random time $n$. Does this phenomena denote that web pages with a higher $\vpi$ probability are more ``valuable" than those with lower $\vpi$ probabilities? For the many of us who use Google daily, it does \cite{anatom:brin1998}. However, for other artifact networks, $\vpi$ can have a completely different meaning.

In journal usage networks, $\vpi$ tends to be a component which makes a distinction between applied and theoretical journals, not ``value" or ``prestige" \cite{mappin:bollen2006}. On the other hand, the $\vpi$ calculated for a journal citation network does provide us with the notion of ``prestige" \cite{journalstatus:bollen2006}. This demonstrates that $\vpi$ has a different meaning depending on the semantics of the edges traversed. In other words, different grammars provide different interpretations of $\vpi$.

Whether $\vpi$ represents ``value" or some other dimension of distinction, this article has provided a method for calculating various $\vpi$ vectors in subsets of the semantic network $G^n$ by means of a random walker algorithm constrained to a grammar. For researchers with nework-based data sets containing heterogeneous entity types and heterogeneous relationship types, this article may provide a more intuitive way of studying the various $\vpi$s of $G^n$.

\section*{Acknowledgments}

This work was funded by a grant from the Andrew W. Mellon Foundation and executed by the MESUR project (http://www.mesur.org).

\end{document}